\documentclass[]{tencent_lightspeed}

\usepackage[utf8]{inputenc}
\usepackage{amsmath}
\usepackage{amsfonts}
\usepackage{amssymb}
\usepackage{booktabs}
\usepackage{tabularx}
\usepackage{multirow}
\usepackage{enumitem}
\usepackage{float}
\usepackage{algorithm}
\usepackage{algorithmic}
\usepackage[most]{tcolorbox}
\usepackage{advdate}
\usepackage{placeins}

\newtcolorbox{placeholderbox}{
  enhanced,
  colback=hunyuansky!4!white,
  colframe=hunyuanblue!70!white,
  boxrule=0.5pt,
  arc=2.5mm,
  left=2mm,
  right=2mm,
  top=1.5mm,
  bottom=1.5mm
}

\definecolor{stevecolor}{HTML}{238A8D}

\title{%
  \raisebox{-0.15em}{%
    \includegraphics[height=1.2em]{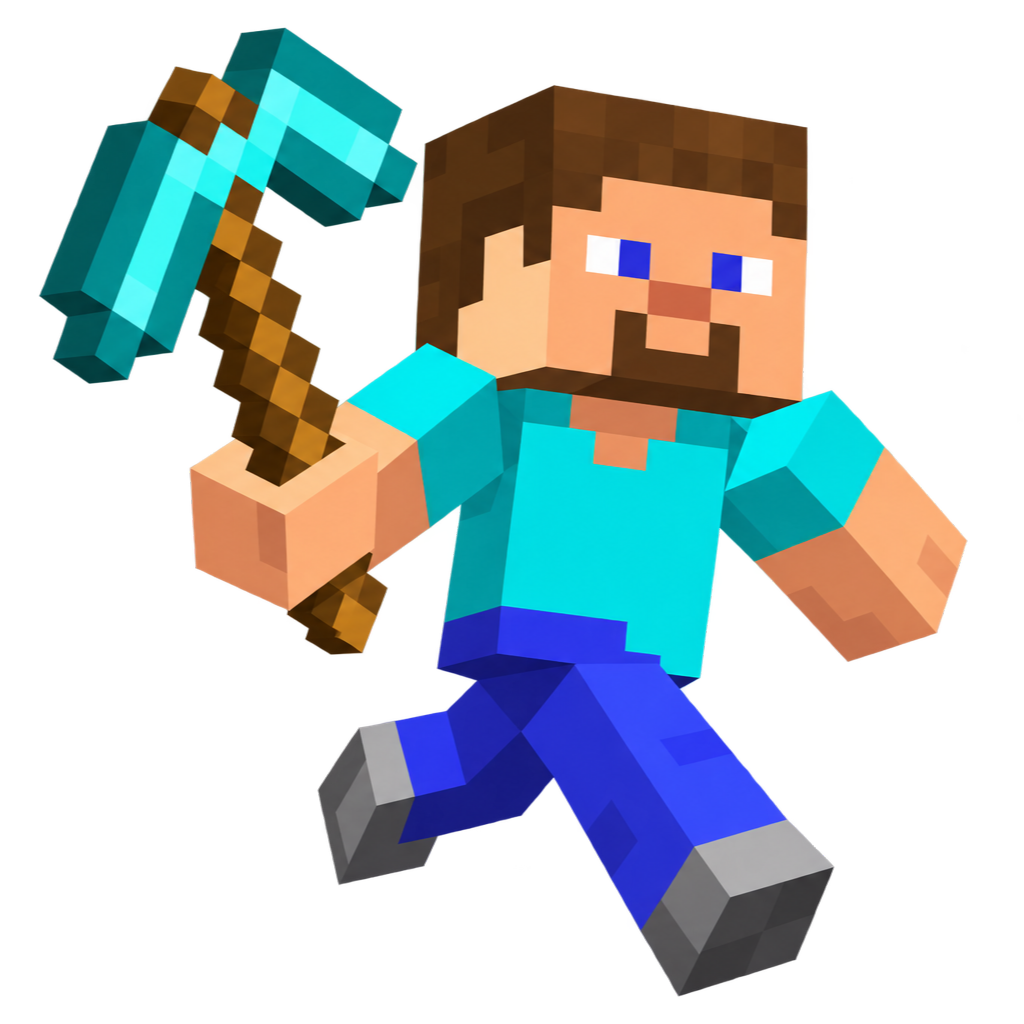}%
  }%
  \hspace{0.15em}%
  {\color{stevecolor}HiSTEVE 1.0:}%
  \hspace{0.22em}%
  Harnessing Human-Interface VLA%
  \\[0.35em]%
  Agents for Open-World Games%
}

\author[1,*]{Zecong Tang}
\author[2,*]{Zian Guan}
\author[3]{Yuncheng Guo}
\author[4,\dagger]{Zhanqiu Zhang}
\author[5,\dagger]{Yiwen Guo}
\author[1,\dagger]{Bo Dai}

\renewcommand{\affiliationlist}{%
  \affiliationformat[1]{The University of Hong Kong},\quad
  \affiliationformat[2]{Zhejiang University},\quad
  \affiliationformat[3]{Fudan University}%
  \\[0.2em]%
  \affiliationformat[4]{LIGHTSPEED},\quad
  \affiliationformat[5]{Independent Researcher}%
}

\contribution[*]{Equal contribution.}
\contribution[\dagger]{Corresponding authors.}

\abstract{
Building virtual embodied agents that operate in open-ended games like human players is a long-standing goal requiring both long-horizon autonomy and reliable native-interface control.
As a procedurally generated open world, Minecraft provides a natural testbed for studying the integration of these capabilities.
Existing agents typically provide either native visuomotor control without long-horizon autonomy or high-level planning executed through environment APIs or predefined skills.
To bridge this gap, we propose HiSTEVE, a hierarchical VLA agent with a capability-aware agent harness.
At the orchestration level, we equip the harness with structured memory for tracking progress and a VLM planner for selecting subtasks within the VLA's capabilities.
To reliably determine subtask completion, we further introduce a policy-coupled task completion state detector that conditions on VLA policy context to signal candidate completion.
At the control level, a single VLA executes all subtasks with vision-only perception and native keyboard-and-mouse actions.
To mitigate interference between heterogeneous GUI and non-GUI interactions, we introduce regime-specific RMSNorm branches and separate mouse-movement token embeddings.
We also train the VLA through world knowledge injection, imitation learning, and reinforcement learning with environment feedback to improve reliability and recovery.
HiSTEVE achieves state-of-the-art atomic-task performance, scoring 85.0\% on MineStudio Bench and a 37.4\% average success rate on OpenHA, while also completing long-horizon tasks such as crafting an iron pickaxe from scratch like human players.
}

\checkdata[Project Page]{\url{https://github.com/Andytonglove/HiSTEVE-1.0}}

\begin{document}
\maketitle
\maintextgeometry

\section{Introduction}
A long-standing goal in embodied intelligence is to build virtual embodied agents \citep{fung2025embodied} that can operate in open-ended games under the same interaction constraints as human players.
Achieving this goal requires two complementary capabilities: long-horizon autonomy to maintain goals and adapt plans; and reliable native-interface control to translate first-person visual observations into keyboard-and-mouse actions.
As a procedurally generated open world, Minecraft provides a natural testbed for studying these capabilities by combining partially observable, compositional tasks with diverse visuomotor interactions \citep{fan2022minedojo,zheng2023mcu}.
A generally capable Minecraft agent must therefore integrate long-horizon planning with native-interface execution in a closed loop, continually updating its decisions from visual feedback.

Existing Minecraft agents have primarily emphasized one of these two capabilities.
On the one hand, visuomotor policies and Vision-Language-Action (VLA) models address native-interface control by directly mapping raw first-person observations to keyboard-and-mouse actions \citep{baker2022video,lifshitz2023steve,li2025jarvis}.
However, these models typically lack mechanisms for persistently tracking task state and explicitly reasoning over task dependencies, limiting their long-horizon autonomy.
On the other hand, agents driven by large language models (LLMs) or vision-language models (VLMs) support long-horizon reasoning through task decomposition planning \citep{wang2023voyager,zhu2023ghost,wang2024jarvis,li2024optimus}.
However, many of these methods still rely on environment APIs, predefined skills, or functional actions for low-level execution, which may be unavailable in most games.
Although recent work has begun to combine high-level planning with learned policies \citep{wang2025openha,park2025mrsteve}, integrating task-state tracking, completion verification, and failure recovery remains an open challenge.

Satisfying both requirements introduces coupled challenges at the planning and action timescales.
At the planning timescale, long-horizon autonomy requires tracking the objective, progress, and execution history across subtasks, while ensuring that plans are consistent with Minecraft mechanics and remain within the lower-level policy's capabilities \citep{wang2023describe}.
The agent must also determine when to terminate the current subtask: terminating too early leaves the subtask unfinished, whereas terminating too late wastes actions and may even undo existing progress.
At the action timescale, native-interface control requires a single VLA policy to operate reliably across heterogeneous task distributions.
In particular, non-GUI control and GUI interaction differ in both visual distributions and mouse-action semantics: mouse movement rotates the camera in the former but displaces the cursor in the latter, making naive joint training susceptible to negative transfer \citep{he2026training,yang2026omniactor}.
Each VLA invocation must also be reliable, since even small per-subtask failure risks accumulate over repeated invocations in long-horizon tasks.

To address these challenges, we propose HiSTEVE, a hierarchical VLA agent that combines a capability-aware harness for long-horizon autonomy and reliable native-interface control.
At the slower planning timescale, we design the harness with a structured memory of task progress and a VLM planner that selects the next atomic subtask.
To close the planning--execution loop, we introduce a policy-coupled task completion state detector that conditions on VLA policy context to signal candidate completion.
The harness then verifies the task state and decides whether to advance, recover, or replan.
At the faster action timescale, the VLA predicts native keyboard-and-mouse actions from the subtask instruction and first-person visual observation.
To mitigate negative transfer across heterogeneous interaction regimes, we introduce regime-specific RMSNorm branches and separate mouse-movement token embeddings for non-GUI and GUI tasks.
Following world knowledge injection and imitation learning, we further finetune the VLA through reinforcement learning (RL) to improve recovery when execution deviates from demonstration trajectories.

We comprehensively evaluate HiSTEVE on atomic and long-horizon benchmarks.
Compared with the strongest baseline in our main comparison, HiSTEVE improves average success by $20.0$ percentage points on MineStudio Bench and $0.6$ points on OpenHA, with family-level gains of up to $28.6$ points on LongBench.
The largest long-horizon improvements occur in tasks requiring multi-stage resource production or the reuse of information gathered during exploration.
Ablations reveal that stronger atomic execution alone does not yield comparable gains in long-horizon performance.
With the VLA held fixed, the full harness more than doubles goal success relative to fixed-horizon handoff, highlighting the importance of organizing local execution around a persistent task state.
The learned completion detector also outperforms prompted general-purpose VLMs, providing more accurate completion judgments to support this execution loop.

The main contributions of this work are summarized as follows:
\begin{itemize}
\item We present HiSTEVE, a hierarchical VLA agent with a capability-aware agent harness, enabling HiSTEVE to maintain goals and adapt plans throughout long-horizon execution.

\item We enable reliable native-interface control through regime-specific RMSNorm branches, separate mouse-movement token embeddings, and a three-stage training strategy.

\item HiSTEVE achieves state-of-the-art atomic-task performance and can complete long-horizon tasks such as crafting an iron pickaxe from scratch like a human player.
\end{itemize}

\section{Related Work}

\subsection{Agent Harnesses for Execution}

Foundation-model agents often use an agent harness, an external orchestration layer that supports interaction across model calls \citep{sumers2023cognitive,wang2025openhands,lin2026agentic}.
Early language agents connect language models to external knowledge and tools and alternate between reasoning and action \citep{nakano2021webgpt,karpas2022mrkl,yao2022react}.
Later systems added memory, feedback-based reflection, and search-based evaluation, allowing the harness to retain experience and refine subsequent decisions \citep{zhou2023language,packer2023memgpt,shinn2023reflexion,wang2024agent,xu2026mem}.
In long-horizon embodied tasks, agent harnesses coordinate high-level planning and low-level execution across multiple steps \citep{huang2022language,wang2024mosaic,shah2025bumble}.
Such harnesses use environmental observations, feasibility assessments, and execution outcomes to continually update subsequent decisions, thereby closing the planning--execution loop \citep{huang2022inner,guo2024doremi,zhi2025closed}.

\subsection{VLA Models and Systems}

VLA models map visual observations and language instructions to embodied actions \citep{ma2026survey,brohan2023rt}.
Early language-conditioned policies learned this mapping directly from multi-task demonstrations \citep{shridhar2022cliport,jang2022bc,brohan2022rt}.
As model capacity and robot datasets grew, these policies generalized to a broader range of tasks, environments, and robot embodiments \citep{brohan2023rt,o2024open,team2024octo,kim2024openvla,black2026pi0visionlanguageactionflowmodel,pertsch2025fast}.
To support long-horizon planning, hierarchical systems use a semantic planner to generate policy-compatible subtasks for execution by a low-level policy \citep{ma2026survey,ahn2022can,driess2023palm,shi2025hi}.
More recent approaches allow intermediate reasoning or predictive representations to directly guide low-level control \citep{belkhale2024rt,yang2025lohovla,lin2025onetwovla,zhang2026pelican}.

\subsection{Game-Playing Agents in Virtual Worlds}

Games have long served as testbeds for sequential decision making, with research progressing from search-based systems to deep RL agents
\citep{campbell2002deep,mnih2015human,silver2016mastering}.
More recently, pretrained models have been used to build agents that follow instructions and reason over observations \citep{ma2024large,zhou2026main,tan2025lumine}.
Researchers have also trained unified policies on multi-game data, but such policies often sacrifice control precision and long-horizon reliability \citep{raad2024scaling,magne2026nitrogen,zhang2026towards}.
Alongside efforts toward cross-game generalization, Minecraft has emerged as a focused testbed for studying embodied interaction within an open world \citep{guss2019minerl,fan2022minedojo}.
Research in Minecraft includes visuomotor policies learned directly from interaction trajectories \citep{baker2022video,lifshitz2023steve,li2025jarvis}, as well as LLM- or VLM-based agents for autonomous exploration and skill acquisition \citep{zhu2023ghost,wang2023voyager,wang2024jarvis}.

\section{Capability-Aware Harness for Long-Horizon Autonomy}
\label{sec:methodology}

We propose a capability-aware agent harness that coordinates a VLM planner and a VLA policy for long-horizon autonomy.
Specifically, we integrate capability-constrained planning, evidence-driven handoff and recovery, and structured memory (Appendix~\ref{app:harness}).

\subsection{Capability-Constrained Planning}
Long-horizon Minecraft tasks cannot be reliably solved through a single static plan because the environment is partially observable, and local execution failures may change the feasibility of subsequent subtasks.
Planning within the harness is reference-free: no predefined subtask sequence or expert task route is provided, and the planner selects the next atomic subtask online from the final objective and current execution context.
We constrain the planner to select and instantiate a parameterized atomic subtask schema from a set restricted to the VLA's execution capabilities.
Each schema specifies its applicability, semantic parameters, visually observable progress and completion conditions, recognizable failure conditions, and execution budget. 

Let $G$ denote the long-horizon objective, $O_j$ the current and recent visual observations at the $j$-th decision step, and $h_j$ the structured planning state, which summarizes verified task progress, unresolved requirements, and previous execution outcomes.
At each decision step, the harness evaluates schema applicability against $h_j$ to construct the currently applicable schema set $\mathcal{S}_j$, while $R_j$ contains contextual memory retrieved according to the current goal and planning state.
Given the planning context and admissible schema set constructed by the harness, the VLM planner selects and instantiates the atomic subtask as
\begin{equation}
u_j
=
\mathcal{P}_{\mathrm{VLM}}
\left(
G,
O_j,
h_j,
\mathcal{S}_j,
R_j
\right),
\label{eq:subtask_selection}
\end{equation}
The structured output $u_j$ specifies a semantic objective rather than a callable skill or scripted action sequence.
The harness converts it into a fixed natural-language instruction $I_j$, which the same VLA executes through native keyboard-and-mouse actions within a bounded invocation.

\begin{figure}[t]
    \centering
    \includegraphics[width=\linewidth]{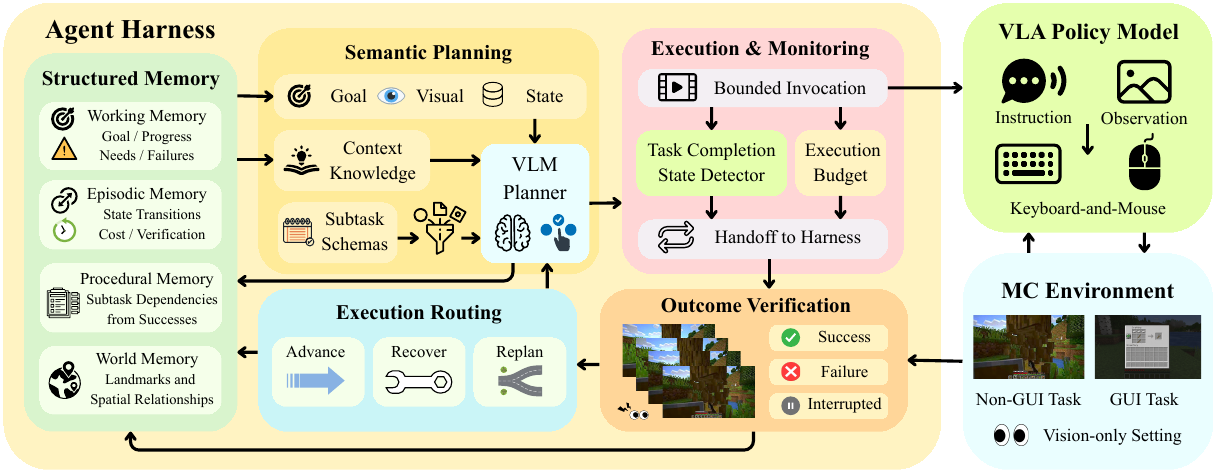}
    \caption{System overview of HiSTEVE, illustrating the agent harness mechanism.}
    \label{fig:system-overview}
\end{figure}

\subsection{Evidence-Driven Handoff and Recovery}

To address the handoff-timing problem between high-level planning and VLA control discussed above, we introduce an evidence-driven handoff-and-recovery protocol.
During each bounded VLA invocation, the task-completion state detector estimates a completion probability for the current instruction $I_j$.
When this probability exceeds a confidence threshold, the current invocation ends and returns control to the harness, and exhaustion of the execution budget triggers the same handoff.
Our mechanism differs from failure detection: the detector signals candidate completion rather than execution failure, while the harness verifies progress and completion conditions using current and recent visual observations.
Based on this reassessment, the execution outcome $v_j$ is recorded as $\mathrm{success}$, $\mathrm{failure}$, or $\mathrm{interrupted}$, where $\mathrm{interrupted}$ indicates that control has been returned before either success or failure can be verified.
Following verification, the harness advances if the completion conditions are satisfied and attempts local recovery if the subtask remains incomplete but recoverable.
The harness replans from the updated state if recovery is infeasible or repeatedly fails, the current subtask becomes inapplicable, or the visual state changes substantially.

\subsection{Structured Memory}

Long-horizon execution requires persistent task state, but retaining the full history increases context length and accumulates unverified intermediate information.
The harness therefore maintains a compact structured memory centered on verified state transitions.
The memory system comprises four stores with distinct roles.
Working memory corresponds to the current planning state $h_j$ and records the active objective and atomic subtask, verified task progress, unresolved requirements, and recent failures or lack of progress.
Episodic memory records the planning states before and after each VLA invocation, together with the selected subtask, the harness-verified outcome, execution cost, and diagnostic information.
Procedural memory stores reusable dependencies among parameterized subtasks derived from verified successful executions.
World memory stores landmarks and spatial relations inferred from visual observations to support subsequent planning within the current world.
At each high-level decision step, the working memory $h_j$ is used directly, whereas task-relevant entries are retrieved from the remaining memory categories to form the $R_j$.

\section{VLA Policy for Native Visuomotor Control}

The lower-level VLA policy maps a subtask instruction with first-person visual observations to native keyboard-and-mouse actions.
We describe the policy architecture and action representation, followed by a three-stage training strategy for improving execution reliability.

\subsection{VLA Policy Architecture}

\paragraph{Architecture.}
We initialize the policy from Qwen2.5-VL-7B \citep{bai2025qwen25vltechnicalreport}.
At control step $t$, the policy takes $I_j$ and the current first-person observation $o_t$ as input and autoregressively generates the tokenized action $\mathbf{a}_t$.
After executing $\mathbf{a}_t$, the agent receives a new observation and repeats this process at 20 Hz.
Non-GUI and GUI interactions require different visuomotor skills, which can lead to conflicting optimization signals when all parameters are shared.
Gradient analysis of a fully shared baseline shows that cross-regime gradients are less aligned than within-regime gradients, with the two RMSNorm modules exhibiting the strongest misalignment among shared components.
The complete measurement protocol and results analysis are provided in Appendix~\ref{app:gradient_conflict}.

Based on this analysis, we introduce Regime-Specific RMSNorm (RS-Norm), a novel lightweight normalization mechanism for heterogeneous interaction regimes.
Specifically, for both the pre-attention and pre-MLP RMSNorms in each decoder layer, we maintain two learnable scale vectors, one for non-GUI interactions and one for GUI interactions.
The appropriate scale vector is selected according to the interaction regime of the current subtask.
For interaction type $c\in\{\mathrm{NG},\mathrm{GUI}\}$, the
RMSNorm at decoder layer $\ell$ and module
$q\in\{\mathrm{attn},\mathrm{mlp}\}$ is defined as
\begin{equation}
\operatorname{RMSNorm}_{\ell,q}^{(c)}(\mathbf h)
=
\boldsymbol{\gamma}_{\ell,q}^{(c)}
\odot
\frac{\mathbf h}
{\sqrt{\frac{1}{d}\sum_{i=1}^{d}h_i^2+\epsilon}},
\label{eq:branched-rmsnorm}
\end{equation}
where $\boldsymbol{\gamma}_{\ell,q}^{(c)}\in\mathbb R^d$ is the
interaction-specific scale vector, while all other parameters remain shared.
For a model with $L$ decoder layers and hidden size $d$, this design adds only $2Ld$ parameters relative to the original backbone, resulting in minimal parameter overhead.
RS-Norm can also be extended to other heterogeneous task combinations, such as vision-language understanding and visuomotor control, as well as multimodal understanding and generation.

\begin{figure}[t]
    \centering
    \includegraphics[width=\linewidth]{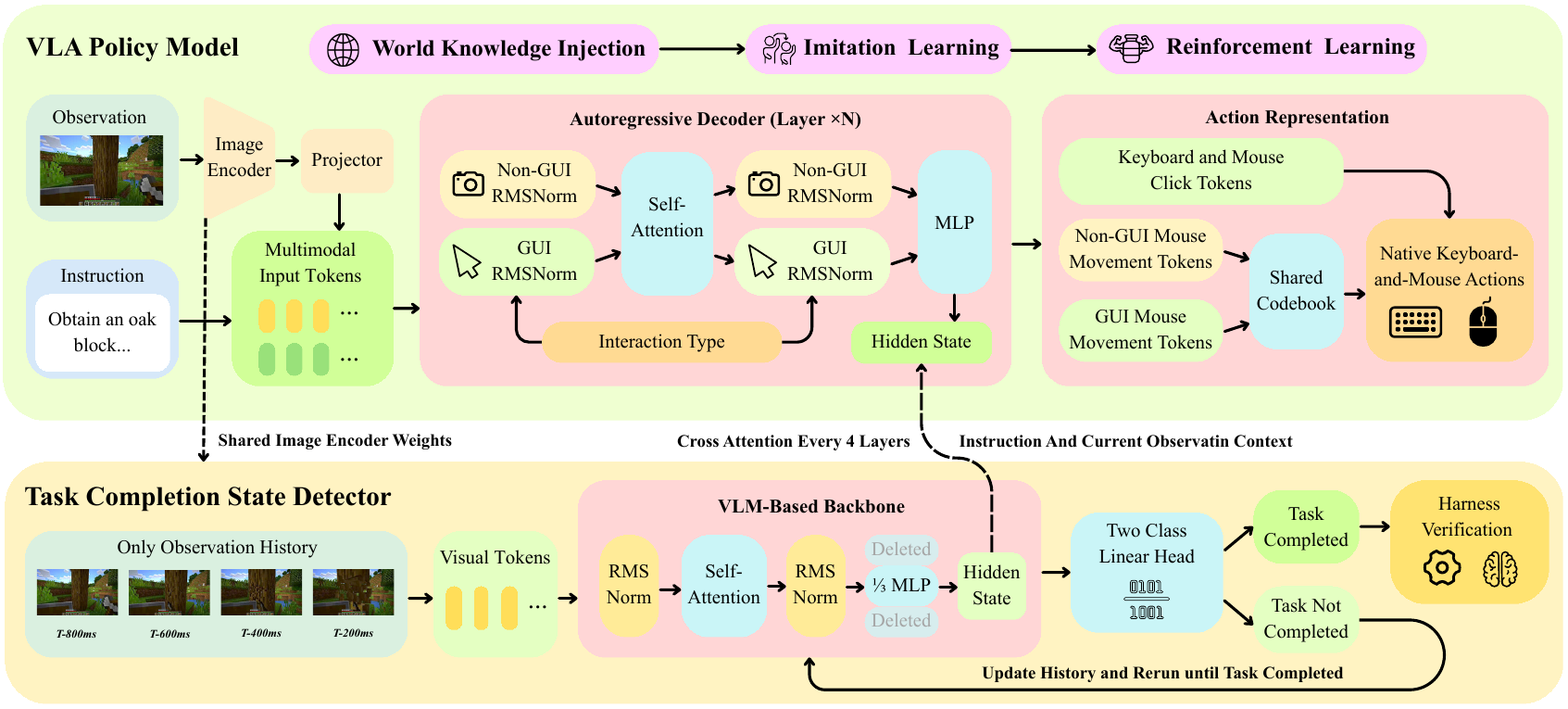}
    \caption{
        Architecture of the VLA policy model and task completion state detector.
    }
    \label{fig:vla-detector}
\end{figure}

\paragraph{Action Representation.}
We represent Minecraft's native keyboard-and-mouse actions as discrete action tokens, allowing supervised action prediction to use the same next-token prediction objective as language modeling.
Following the VPT-style action space \citep{baker2022video}, we organize each single-step action into multiple categorical action groups.
Keyboard and click inputs are represented as discrete choices within predefined action groups, while horizontal and vertical mouse displacements are $\mu$-law transformed and quantized into the same number of bins.

However, mouse displacement controls camera rotation in non-GUI interactions but cursor movement in GUI interactions, so sharing a single action token set creates an ambiguity.
We therefore assign disjoint mouse-displacement token sets: the original mouse tokens are retained for non-GUI interactions, while a separate subset is allocated to GUI interactions.
We further insert an interaction-type token at the beginning of the sequence to indicate whether the current interaction is non-GUI or GUI.
Because both sets of mouse-displacement tokens decode back to the same physical action codebook, the branching occurs solely at the representation level. We refer to this regime-separated mouse-movement token representation as \textbf{Dual-Mouse}. (Appendix~\ref{app:action_representation})

\subsection{Three-Stage Policy Training}
\label{sec:three_stage}

Because per-subtask failures compound over repeated VLA invocations, reliable long-horizon execution requires high reliability. We therefore train the VLA in three stages: world knowledge injection, imitation learning, and reinforcement learning with environment feedback.

\paragraph{Supervised Initialization.}
Before reinforcement learning, we use two supervised stages to establish the semantic knowledge and control capabilities required for reliable native-interface execution.
World knowledge injection equips the policy with Minecraft concepts and their visual grounding, while imitation learning teaches it to map subtask instructions and first-person observations to native keyboard-and-mouse actions.
Further details on data sources, sample construction and curation, training objectives, and optimization settings are provided in Appendix~\ref{app:vla_training}.

\paragraph{Reinforcement Learning}
After imitation learning, the policy remains less reliable in GUI interactions, because precise cursor positioning and multi-step operations make GUI execution more sensitive to small action errors.
We address this imbalance through environment-verified branch comparisons at both trajectory and state-matched segment levels.
For each initial condition $s_0$, we independently sample $K$ trajectories using a behavior policy $\pi_b$.
Branch outcomes can be scored using either VLMs judge or direct environment feedback. Considering cost and efficiency, we adopt the latter for trainer-side reward construction and state matching.
Each trajectory score $R_i$ combines certified completion, maximum verified progress, execution cost, and failure penalties. The resulting trajectories form a same-start branch group, whose scores yield the group-relative advantage:
\begin{equation}
\mathcal{B}(s_0)=\{\tau_i\}_{i=1}^{K},
\quad
\tau_i\sim\pi_b(\cdot\mid s_0),
\qquad
A_i=
\frac{R_i-\mu_{\mathcal{B}}}
{\sigma_{\mathcal{B}}+\varepsilon_A},
\label{eq:branch_group_advantage}
\end{equation}
where $\mu_{\mathcal{B}}$ and $\sigma_{\mathcal{B}}$ denote the within-group score mean and standard deviation.

We optimize informative branch groups with clipped GRPO \citep{shao2024deepseekmathpushinglimitsmathematical} updates, separating detached behavior-policy importance correction from the current-policy update ratio.
However, trajectory-level advantages can penalize correct prefixes in failed rollouts.
We therefore construct local preference pairs from successful and failed branches, prioritizing the same rollout origin and matching structured signatures of inventory, GUI, cursor, execution stage, and environment events.
The resulting DPO-style pairwise loss targets divergent action segments, using the same completion certificates as group-relative learning.
To preserve non-GUI capabilities, we interleave branch updates with world-action replay and regularize against a fixed reference policy $\pi_{\mathrm{ref}}$:
\begin{equation}
\mathcal{L}_{\mathrm{post}}
=
\mathcal{L}_{\mathrm{branch}}
+
\lambda_{\mathrm{world}}
\mathcal{L}_{\mathrm{world}}
+
\lambda_{\mathrm{ref}}
D_{\mathrm{KL}}
\left(
\pi_\theta
\Vert
\pi_{\mathrm{ref}}
\right),
\label{eq:post_training_objective}
\end{equation}
where $\mathcal{L}_{\mathrm{branch}}$ is the group-relative loss for informative branch groups or the pairwise loss for matched segments, depending on the available comparisons, and $\mathcal{L}_{\mathrm{world}}$ is the world-action replay loss.
Reward construction and optimization details are provided in Appendix~\ref{app:rl_details}.

\section{Task Completion State Detector}

We introduce a novel policy-coupled task completion state detector to support the harness in determining when to pause the VLA invocation.
To reduce system overhead, the detector shares the VLA image encoder and additionally conditions on VLA policy context for completion prediction, with Qwen2.5-VL-3B serving as its backbone.
At each query, the detector uses the shared image encoder and its self-attention stream to process historical frames sampled at $800$, $600$, $400$, and $200\,\mathrm{ms}$ before the query, while the VLA processes the current frame and subtask instruction.
The final-layer hidden states of the input prefix preceding the first action token form the policy context $H_t^\pi$, which is injected through cross-attention after every fourth detector decoder layer.
Let $Z_t^{(\ell)}$ denote the detector hidden states at layer $\ell$. We apply query-normalized residual cross-attention as
\begin{equation}
\widetilde{Z}_t^{(\ell)}
=
Z_t^{(\ell)}
+
\operatorname{MHA}_{\ell}
\left(
\operatorname{LN}_{\ell}(Z_t^{(\ell)}),
H_t^\pi,
H_t^\pi
\right),
\label{eq:detector-cross-attention}
\end{equation}
where each cross-attention module has independent projection parameters.
To further reduce detector overhead, we use Wanda-inspired activation--weight channel scores to prune the intermediate width of each feed-forward network to approximately one third \citep{sun2024simple}.
For classification, we map the detector's final-token hidden state to two logits through a linear classification head and apply a softmax to obtain the completion probability $p_t$.
We train the detector with the VLA frozen on completion labels derived from VPT human gameplay trajectories \citep{baker2022video}.
At deployment, the detector runs before action decoding and raises a candidate completion signal to the harness only when $p_t$ exceeds $\tau$ for two consecutive queries.
Details of dataset construction, channel pruning and training are provided in Appendix~\ref{app:detector_details}.

\section{Experiments}
\label{sec:experiments}

\subsection{Experimental Setup}
\label{sec:experimental_setup}

We instantiate HiSTEVE with GPT-4o~\citep{hurst2024gpt} as the upper-level planner and Qwen2.5-VL-7B~\citep{bai2025qwen25vltechnicalreport} as the low-level VLA.
We evaluate atomic execution on MineStudio Bench~\citep{zheng2023mcu} and OpenHA~\citep{wang2025openha}, and long-horizon agency on LongBench.
Ablation scores are averaged over repeated independent evaluations for each fixed checkpoint.
MineStudio contains $20$ tasks with three trials each, OpenHA reports family-wise episode success over Embodied, GUI, and Combat, and LongBench contains $14$ goal-only tasks with $49$ hidden diagnostic nodes.
We compare against prior native-control agents, including VPT~\citep{baker2022video}, STEVE-1~\citep{lifshitz2023steve}, ROCKET-1~\citep{cai2025rocket}, and JARVIS-VLA~\citep{li2025jarvis}, as well as OpenHA variants---LatentHA, MotionHA, GroundingHA, TextVLA, and Text/CoA~\citep{wang2025openha}.
Complete evaluation protocols are provided in Appendix~\ref{app:evaluation_protocol}.

\subsection{Main Results}
\label{sec:main_results}

As shown in Table~\ref{tab:main_results}, we evaluate HiSTEVE across atomic and long-horizon tasks.

\begin{table*}[t]
\centering
\scriptsize
\caption{
Comparison on MineStudio Bench, OpenHA, and LongBench.
MineStudio reports strict pass@1, OpenHA reports family-wise episode success and macro average, and LongBench reports final-goal success.
R\&C, E\&M, Rec., and Open denote the four LongBench task families.
The best and second-best results are highlighted in \textbf{bold} and \underline{underline}, respectively.
}
\label{tab:main_results}
\setlength{\tabcolsep}{2.6pt}
\renewcommand{\arraystretch}{1.08}
\resizebox{\textwidth}{!}{%
\begin{tabular}{l|cccc|cccc|cccc}
\toprule
& \multicolumn{4}{c|}{\textbf{MineStudio Bench}}
& \multicolumn{4}{c|}{\textbf{OpenHA}}
& \multicolumn{4}{c}{\textbf{LongBench}} \\
\cmidrule(lr){2-5}
\cmidrule(lr){6-9}
\cmidrule(lr){10-13}
\textbf{Method}
& \textbf{Mine}
& \textbf{Kill}
& \textbf{Craft}
& \textbf{Smelt}
& \textbf{Emb.}
& \textbf{GUI}
& \textbf{Combat}
& \textbf{Avg.}
& \textbf{R\&C}
& \textbf{E\&M}
& \textbf{Rec.}
& \textbf{Open} \\
\midrule

VPT
& 20.0 & 13.3 & 0.0 & 6.7
& 10.1 & 0.7 & 3.6 & 4.8
& 0.0 & 1.4 & 0.0 & 0.0 \\

STEVE-1
& 20.0 & 13.3 & 0.0 & 6.7
& 8.4 & 0.0 & 4.9 & 4.4
& 0.0 & 0.7 & 0.0 & 0.0 \\

ROCKET-1
& 33.3 & 40.0 & 0.0 & 6.7
& 19.2 & 0.0 & 29.8 & 16.3
& 1.4 & 2.9 & 0.7 & 0.0 \\

JARVIS-VLA
& 66.7 & 60.0 & 26.7 & 60.0
& 31.0 & 25.3 & 18.3 & 24.9
& 7.1 & 5.7 & 1.4 & 0.7 \\

LatentHA
& 46.7 & 33.3 & 6.7 & 20.0
& 27.3 & 3.5 & 8.2 & 13.0
& 2.1 & 2.9 & 0.7 & 0.0 \\

MotionHA
& 53.3 & 33.3 & 6.7 & 20.0
& 31.6 & 0.0 & 9.1 & 13.6
& 2.9 & 3.6 & 0.7 & 0.0 \\

GroundingHA
& 60.0 & 53.3 & 13.3 & 33.3
& \underline{39.7} & 3.7 & 28.2 & 23.9
& 5.7 & 4.3 & 1.4 & 0.0 \\

TextVLA
& 46.7 & 46.7 & 13.3 & 40.0
& 23.9 & 14.0 & 27.1 & 21.7
& 7.1 & 5.0 & 1.4 & 0.7 \\

OpenHA Text/CoA
& \underline{73.3}
& \underline{73.3}
& \underline{40.0}
& \underline{73.3}
& 37.0
& \textbf{33.3}
& \textbf{40.0}
& \underline{36.8}
& \underline{14.3}
& \underline{7.1}
& \underline{2.9}
& \underline{1.4} \\

\midrule
\textbf{HiSTEVE (Ours)}
& \textbf{93.3}
& \textbf{86.7}
& \textbf{66.7}
& \textbf{93.3}
& \textbf{46.0}
& \underline{27.2}
& \underline{39.0}
& \textbf{37.4}
& \textbf{42.9}
& \textbf{28.6}
& \textbf{14.3}
& \textbf{14.3} \\

\bottomrule
\end{tabular}%
}
\end{table*}

\paragraph{Atomic task execution.}

On MineStudio Bench, HiSTEVE achieves strict pass@1 scores of $93.3\%$, $86.7\%$, $66.7\%$, and $93.3\%$ on Mine, Kill, Craft, and Smelt, respectively, corresponding to an overall average of $85.0\%$.
Compared with OpenHA Text/CoA, HiSTEVE improves the four-family average from $65.0\%$ to $85.0\%$, with gains of $20.0$, $13.4$, $26.7$, and $20.0$ percentage points on Mine, Kill, Craft, and Smelt, respectively. The largest absolute gain occurs on Craft, although Craft remains the weakest MineStudio category, indicating that multi-step GUI manipulation remains more challenging than first-person mining and combat.

On OpenHA, HiSTEVE achieves a $37.4\%$ average success rate, compared with $36.8\%$ for OpenHA Text/CoA. The gain is concentrated in Embodied tasks, where success increases from $37.0\%$ to $46.0\%$, while GUI and Combat remain slightly below the corresponding baseline scores.
Our trajectory analysis shows that failed GUI episodes can reach and interact with the correct interface but break down during multi-step item placement, intermediate-state correction, or result retrieval, indicating that robust multi-step GUI execution remains a key bottleneck.

\paragraph{Long-horizon execution.}

LongBench evaluates whether atomic native-control capabilities remain effective when they must be repeatedly composed over extended interaction.
Unlike atomic benchmarks, success depends not only on executing individual subtasks, but also on maintaining task state across invocations and adapting subsequent decisions to observed outcomes.
We evaluate Resource \& Crafting, Exploration \& Memory, Recovery, and Open-ended tasks using final-goal success as the primary metric, with hidden diagnostic nodes used only for progress analysis.

\begin{figure}[t]
\centering
\includegraphics[width=\linewidth]{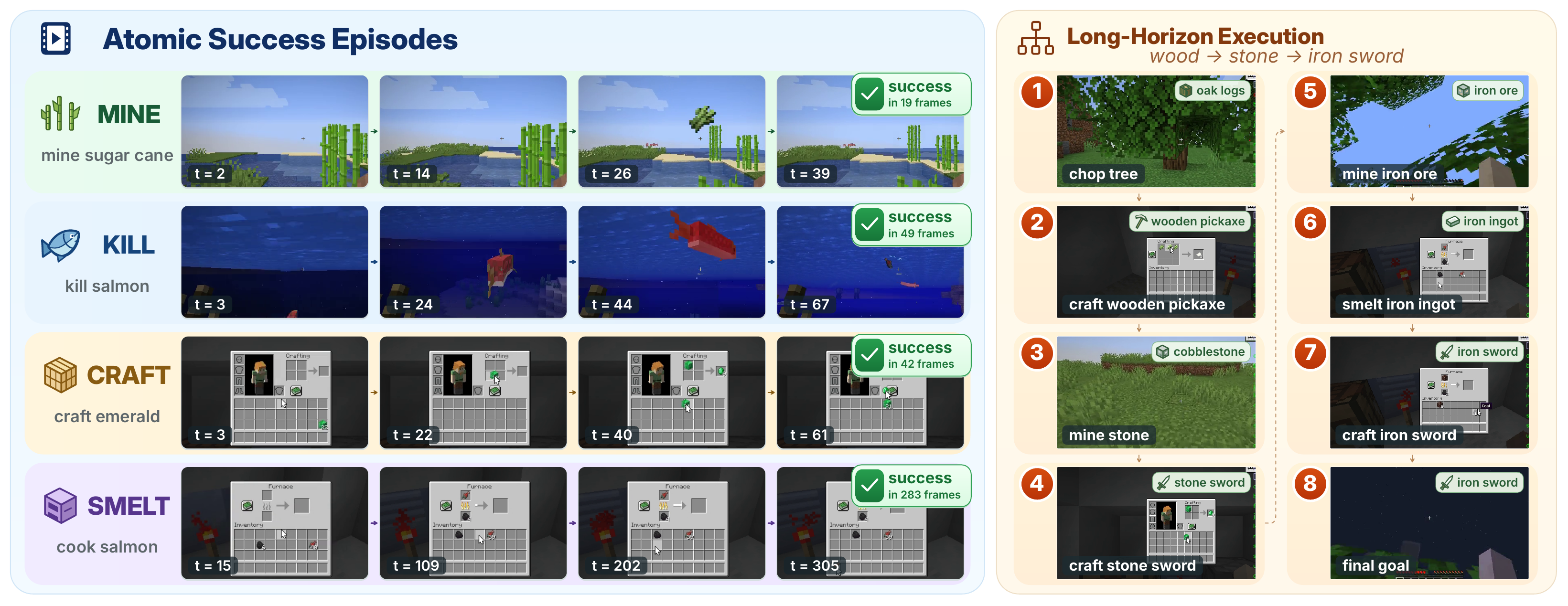}
\caption{Qualitative examples of HiSTEVE. The left panel shows successful atomic-task executions, while the right panel illustrates multi-stage long-horizon execution.}
\label{fig:success_qual}
\end{figure}

HiSTEVE achieves final-goal success rates of $42.9\%$, $28.6\%$, $14.3\%$, and $14.3\%$ on Resource \& Crafting, Exploration \& Memory, Recovery, and Open-ended tasks, respectively, with the overall Goal SR in Table~\ref{tab:main_ablation} computed as the macro-average over the four task families.
Relative to OpenHA Text/CoA, these correspond to improvements of $28.6$, $21.5$, $11.4$, and $12.9$ percentage points.
The largest gains appear on Resource \& Crafting and Exploration \& Memory, where successful execution requires repeated composition of atomic skills and persistent task-state tracking.
Recovery and Open-ended tasks remain substantially harder, indicating that robust recovery and weakly structured objectives remain important challenges at longer horizons.
Figure~\ref{fig:success_qual} complements these results with qualitative examples of atomic and long-horizon execution.

\subsection{Atomic Task Completion Detection Evaluation}
\label{sec:completion_detection_eval}

\begin{table}[t]
\centering
\setlength{\tabcolsep}{6pt}
\renewcommand{\arraystretch}{1.15}
\caption{
Atomic-subtask completion detection. All models use the same temporal sampling; the random baseline has an expectation of $0.5$ on every metric under the balanced validation set.
}
\label{tab:state_detector}
\begin{tabular}{l r cccccc}
\toprule
Method & \#Params & Acc $\uparrow$ & Prec $\uparrow$ & Rec $\uparrow$ & FPR $\downarrow$ & FNR $\downarrow$ & F1 $\uparrow$ \\
\midrule
Random   & --            & 0.500 & 0.500 & 0.500 & 0.500 & 0.500 & 0.500 \\
Qwen3.5  & 9B            & 0.500 & 0.504 & 0.030 & \textbf{0.030} & 0.970 & 0.057 \\
Qwen3.5  & 35B           & 0.502 & 0.530 & 0.042 & 0.037 & 0.958 & 0.078 \\
GPT-4o   & --            & 0.698 & 0.708 & 0.673 & 0.277 & 0.327 & 0.690 \\
GPT-5.5  & --            & 0.780 & 0.791 & 0.761 & 0.201 & 0.239 & 0.776 \\
\midrule
\textbf{Ours} & 1.4B     & \textbf{0.863} & \textbf{0.861} & \textbf{0.865} & 0.140 & \textbf{0.135} & \textbf{0.863} \\
\bottomrule
\end{tabular}
\end{table}


We evaluate completion detection on a class-balanced validation split of $4{,}674$ samples drawn from VPT trajectories~\citep{baker2022video} across four task families: mining, killing, crafting, and smelting (details see Appendix~\ref{app:detector_dataset}).
As baselines, we use a random classifier, the open-source Qwen3.5 VLMs with 9B and 35B parameters~\citep{qwen35blog}, and the closed-source GPT-4o and GPT-5.5 models~\citep{hurst2024gpt,singh2025openai}. These general-purpose models are not trained for task-completion detection. We provide each model with the same history frames, current frame, and subtask instruction, and prompt it to make a binary completion decision.
As reported in Table~\ref{tab:state_detector}, we measure overall accuracy, precision, recall, and F1 for the completion class. We additionally report the false-positive rate (FPR) and false-negative rate (FNR), which quantify premature completion predictions and missed completions, respectively.

As shown in Table~\ref{tab:state_detector}, our specialized detector consistently outperforms the prompted general-purpose VLM baselines.
Qwen3.5 strongly favors the incomplete class, with recalls of only $0.030$ and $0.042$ at 9B and 35B parameters, respectively.
GPT-4o and GPT-5.5 improve F1 to $0.690$ and $0.776$.
In contrast, our detector reaches $0.863$ in both accuracy and F1, with precision and recall of $0.861$ and $0.865$, providing a more accurate and balanced candidate-completion signal for the planning--control interface.
The downstream effect of state-dependent handoff within the complete agent is evaluated in Section~\ref{sec:ablation}, while detector architectural ablations are reported in Appendix~\ref{app:detector_ablation}.

\subsection{Ablation Study}
\label{sec:ablation}

Table~\ref{tab:main_ablation} isolates staged policy training, regime-specific representation, and harness orchestration. 

\begin{table*}[t]
\centering
\scriptsize
\caption{
Ablation of staged VLA training, regime-specific representation, and agent harness design.
For MineStudio, World averages Mine and Kill, GUI averages Craft and Smelt.
For OpenHA, World averages Embodied and Combat, GUI denotes the GUI family.
LongBench reports final-goal success (Goal SR) and hidden diagnostic progress (Progress); Full H. denotes the complete agent harness.
}
\label{tab:main_ablation}

\setlength{\tabcolsep}{2.8pt}
\renewcommand{\arraystretch}{1.10}
\resizebox{\textwidth}{!}{%
\begin{tabular}{lccccc|ccc|ccc|cc}
\toprule
& \multicolumn{5}{c|}{\textbf{Method components}}
& \multicolumn{3}{c|}{\textbf{MineStudio}}
& \multicolumn{3}{c|}{\textbf{OpenHA}}
& \multicolumn{2}{c}{\textbf{LongBench}} \\
\cmidrule(lr){2-6}
\cmidrule(lr){7-9}
\cmidrule(lr){10-12}
\cmidrule(lr){13-14}

\textbf{Method}
& \textbf{WKI}
& \textbf{RL}
& \textbf{RS}
& \textbf{DM}
& \textbf{Full H.}
& \textbf{World}
& \textbf{GUI}
& \textbf{All}
& \textbf{World}
& \textbf{GUI}
& \textbf{Avg.}
& \textbf{Goal SR}
& \textbf{Progress} \\
\midrule

\multicolumn{14}{l}{\textit{Staged policy training}} \\

Imitation learning only
& -- & -- & $\checkmark$ & $\checkmark$ & --
& 54.7 & 46.2 & 50.4
& 36.4 & 15.7 & 29.5
& 7.1 & 11.9 \\

\quad + world knowledge injection
& $\checkmark$ & -- & $\checkmark$ & $\checkmark$ & --
& 71.4 & 58.9 & 65.1
& 41.8 & 22.4 & 35.3
& \underline{14.3} & 20.7 \\

\quad + environment-feedback RL
& $\checkmark$ & $\checkmark$ & $\checkmark$ & $\checkmark$ & --
& 89.7 & 71.6 & 80.7
& \underline{43.1} & \underline{26.8} & \underline{37.7}
& \underline{14.3} & \underline{21.9} \\

\midrule
\multicolumn{14}{l}{\textit{Regime-specific representation}} \\

Shared representation
& $\checkmark$ & $\checkmark$ & -- & -- & --
& 47.8 & 34.5 & 41.1
& 34.2 & 11.5 & 26.6
& 3.6 & 8.6 \\

\quad + RS-Norm
& $\checkmark$ & $\checkmark$ & $\checkmark$ & -- & --
& 62.9 & 51.8 & 57.3
& 36.9 & 19.2 & 31.0
& 7.1 & 13.9 \\

\quad + Dual-Mouse
& $\checkmark$ & $\checkmark$ & -- & $\checkmark$ & --
& 69.5 & 54.3 & 61.9
& 39.7 & 18.4 & 32.6
& 10.7 & 16.7 \\

\quad + RS-Norm + Dual-Mouse
& $\checkmark$ & $\checkmark$ & $\checkmark$ & $\checkmark$ & --
& 89.7 & 71.6 & 80.7
& \underline{43.1} & \underline{26.8} & \underline{37.7}
& \underline{14.3} & \underline{21.9} \\

\midrule
\multicolumn{14}{l}{\textit{Harness orchestration}} \\

Full VLA + fixed-horizon handoff
& $\checkmark$ & $\checkmark$ & $\checkmark$ & $\checkmark$ & --
& \textbf{90.5} & \underline{71.9} & \underline{81.2}
& \textbf{43.4} & 26.5 & \textbf{37.8}
& 10.7 & 17.9 \\

\textbf{Full HiSTEVE}
& $\checkmark$ & $\checkmark$ & $\checkmark$ & $\checkmark$ & $\checkmark$
& \underline{90.0} & \textbf{80.0} & \textbf{85.0}
& 42.5 & \textbf{27.2} & 37.4
& \textbf{25.0} & \textbf{28.6} \\

\bottomrule
\end{tabular}%
}
\end{table*}


\paragraph{Effect of regime-specific representation.}

RS-Norm and Dual-Mouse provide complementary gains: from a shared-representation baseline of $41.1\%$ on MineStudio All, they individually improve performance to $57.3\%$ and $61.9\%$, and jointly reach $80.7\%$.
Together, they also improve OpenHA Avg. from $26.6\%$ to $37.7\%$ and LongBench Goal SR from $3.6\%$ to $14.3\%$.
Concurrent gains on both MineStudio World and GUI suggest that regime-specific feature scaling and mouse-action separation mitigate complementary interference between the two interaction regimes.

\paragraph{Effect of staged policy training.}

World knowledge injection substantially improves the lower-level executor, increasing MineStudio All from $50.4\%$ to $65.1\%$ and OpenHA Avg. from $29.5\%$ to $35.3\%$, while LongBench Goal SR increases from $7.1\%$ to $14.3\%$.
Environment-feedback RL further raises MineStudio All to $80.7\%$ and OpenHA Avg. to $37.7\%$, but leaves LongBench Goal SR unchanged at $14.3\%$.
This separation indicates that post-training strengthens local execution reliability, while improving the executor alone is insufficient to solve long-horizon coordination.

\paragraph{Effect of the agent harness.}

With the lower-level VLA fixed, this ablation evaluates system-level execution, where handoff and verification can still affect atomic success.
The complete harness changes atomic performance modestly, from $81.2\%$ to $85.0\%$ on MineStudio All and from $37.8\%$ to $37.4\%$ on OpenHA Avg.
In contrast, LongBench Goal SR increases from $10.7\%$ to $25.0\%$, with diagnostic progress rising from $17.9\%$ to $28.6\%$.
These gains highlight the role of capability-constrained planning and execution handoff and verification in organizing VLA invocations.

\section{Conclusion}
\label{sec:conclusion}

We present HiSTEVE, a hierarchical VLA agent combining a capability-aware agent harness for long-horizon autonomy with native-interface control.
The harness enables long-horizon autonomy through capability-constrained planning, evidence-driven handoff and recovery, and structured memory.
Regime-specific representations and three-stage training enable a single VLA to reliably handle both non-GUI and GUI interactions.
HiSTEVE achieves state-of-the-art performance on MineStudio Bench ($85.0\%$) and OpenHA ($37.4\%$) and can craft an iron pickaxe from scratch using only human-interface.
Our work shows that regime-specific representations support reliable control across heterogeneous tasks, while reliable long-horizon execution further requires a harness that combines local control with state-aware orchestration.
These results represent a step toward virtual embodied agents that operate in open-ended games human players.

\bibliographystyle{iclr2027_conference}
\bibliography{reference}

\beginappendix

\newcommand{\sampleimageplaceholder}[1]{%
    \begingroup
    \setlength{\fboxsep}{0pt}%
    \fbox{%
        \parbox[c][0.18\textwidth][c]{0.32\textwidth}{%
            \centering
            \footnotesize #1\\[3pt]
            \scriptsize $640\times360$
        }%
    }%
    \endgroup
}

\section{Agent Harness Details}
\label{app:harness}

This section provides additional implementation details for the agent harness introduced in Section~\ref{sec:methodology}. We describe the runtime interaction between the VLM planner and the native-control VLA, the capability-constrained atomic interface, evidence-driven verification and recovery, and the structured memory used during long-horizon execution.

\subsection{Harness Runtime}
\label{app:harness_runtime}

HiSTEVE executes a long-horizon objective through repeated bounded interactions between a high-level semantic planner and a continuously acting VLA policy. At high-level decision step $j$, the harness maintains a structured planning state
\begin{equation}
h_j
=
\left(
G,
p_j,
r_j,
v_{<j}
\right),
\label{eq:appendix_planning_state}
\end{equation}
where $G$ denotes the final task objective, $p_j$ denotes visually verified progress, $r_j$ denotes unresolved requirements, and $v_{<j}$ summarizes previous execution outcomes.

Given $h_j$, the harness determines which parameterized atomic subtask schemas are currently applicable and constructs the admissible set
\begin{equation}
\mathcal{S}_j
=
\left\{
s \in \mathcal{S}
\; \middle| \;
\operatorname{Applicable}(s,h_j)=1
\right\}.
\label{eq:appendix_admissible_set}
\end{equation}
Task-relevant information retrieved from episodic, procedural, and world memory is denoted by $R_j$. The planner then selects and instantiates the next atomic subtask:
\begin{equation}
u_j
=
\mathcal{P}_{\mathrm{VLM}}
\left(
G,
O_j,
h_j,
\mathcal{S}_j,
R_j
\right),
\label{eq:appendix_subtask_selection}
\end{equation}
where $O_j$ denotes the current and recent visual observations.

The planner output $u_j$ specifies a semantic objective rather than an executable Minecraft function or scripted action sequence. The harness converts it into a fixed natural-language instruction $I_j$, which is executed by the same VLA during one bounded invocation. At control step $t$ within the invocation, the VLA predicts
\begin{equation}
\mathbf{a}_t
\sim
\pi_{\theta}
\left(
\cdot
\mid
o_t,
I_j
\right),
\label{eq:appendix_vla_runtime}
\end{equation}
where $o_t$ is the current first-person observation and $\mathbf{a}_t$ is a native keyboard-and-mouse action.

The instruction $I_j$ remains fixed throughout the invocation, while the visual observation changes after every environment step. This separation yields two interacting timescales: the VLA performs continuous control at the action timescale, whereas the harness updates the task state only at semantic execution boundaries.

Control is returned to the harness when either the execution budget is exhausted or the completion detector produces a candidate completion signal. Let $p_t$ denote the detector probability that the current subtask has completed. A detector-triggered handoff is produced when
\begin{equation}
p_t > \tau
\quad \land \quad
p_{t-1} > \tau,
\label{eq:appendix_handoff_rule}
\end{equation}
where $\tau$ is the detector operating threshold. Requiring two consecutive positive predictions reduces sensitivity to isolated detector errors.

Importantly, a handoff does not imply successful task completion. It only indicates that the current bounded invocation should stop and return control to the harness for verification.

\subsection{Capability-Constrained Atomic Interface}
\label{app:atomic_interface}

Direct free-form decomposition can produce subtasks that are too abstract for reliable execution by the VLA, or conversely, subtasks that are so narrow that they require unnecessarily frequent high-level decisions. HiSTEVE therefore exposes the lower-level VLA to the planner through a parameterized atomic interface.

Each atomic schema $s \in \mathcal{S}$ is represented as
\begin{equation}
s
=
\left(
\mathcal{A}_s,
\mathcal{P}_s,
\mathcal{C}_s,
\mathcal{F}_s,
B_s
\right),
\label{eq:appendix_schema}
\end{equation}
where $\mathcal{A}_s$ denotes the semantic arguments of the subtask, $\mathcal{P}_s$ its applicability conditions, $\mathcal{C}_s$ its visually observable completion criteria, $\mathcal{F}_s$ recognizable failure conditions, and $B_s$ the execution budget.

The schema library defines the capability boundary presented to the planner. It does not contain scripted implementations of the corresponding subtasks. Once a schema is instantiated, all physical interactions required to realize the semantic objective are produced by the VLA through native keyboard-and-mouse actions.

For count-based objectives, the harness does not issue an indefinitely running instruction such as ``collect $N$ items.'' Instead, it repeatedly invokes bounded acquisition subtasks. After invocation $j$, the remaining requirement is updated according to the visually verified progress:
\begin{equation}
r_{j+1}
=
\max
\left(
0,
r_j-\Delta_j
\right),
\label{eq:appendix_remaining_requirement}
\end{equation}
where $\Delta_j$ denotes newly verified progress during the current invocation.

This decomposition prevents already acquired resources from being repeatedly counted as new progress and avoids continuing a stale local instruction after the corresponding requirement has already been satisfied.

\subsection{Evidence-Driven Verification and Recovery}
\label{app:harness_verification}

HiSTEVE explicitly separates execution handoff from task verification. A detector-triggered handoff only indicates a candidate semantic boundary. After control returns, the harness reassesses the current and recent visual observations and evaluates the completion and failure predicates associated with the active subtask.

The resulting execution outcome is represented as
\begin{equation}
v_j
\in
\left\{
\mathrm{success},
\mathrm{failure},
\mathrm{interrupted}
\right\},
\label{eq:appendix_execution_outcome}
\end{equation}
where $\mathrm{interrupted}$ indicates that control has returned to the harness but neither success nor failure can yet be visually verified.

The updated planning state is then
\begin{equation}
h_{j+1}
=
\mathcal{U}
\left(
h_j,
u_j,
v_j,
O_{j+1}
\right),
\label{eq:appendix_state_update}
\end{equation}
where $\mathcal{U}$ denotes the harness state-update operation and $O_{j+1}$ contains the visual evidence available after the invocation.

During deployment, the planner and VLA interact with the environment only through first-person observations and native keyboard-and-mouse actions. Simulator-side state used during training for reward construction or trajectory certification is not provided to the deployed planner or VLA.

For example, resource acquisition is accepted only after the corresponding inventory change is visually confirmed. Likewise, crafting or smelting is not considered complete merely because an intermediate GUI state is reached; the resulting item must be visually observed in the player inventory.

After verification, the harness follows three possible transitions:
\begin{equation}
\operatorname{Next}(v_j)
=
\begin{cases}
\mathrm{advance}, &
\text{if the completion condition is verified},\\
\mathrm{recover}, &
\text{if the subtask remains locally recoverable},\\
\mathrm{replan}, &
\text{otherwise}.
\end{cases}
\label{eq:appendix_harness_transition}
\end{equation}

Recovery is used for local execution deviations that do not invalidate the high-level objective, such as temporary lack of progress, inconsistent GUI state, target loss, or a produced item that has not yet been collected. If recovery is infeasible or repeatedly fails, the current subtask becomes inapplicable, or the observed state changes substantially, the harness invokes the planner again from the updated task state.

\subsection{Structured Memory}
\label{app:structured_memory}

Long-horizon execution requires persistent task state, but retaining the complete raw interaction history rapidly increases context length and preserves many transient or unverified intermediate states. HiSTEVE instead maintains a compact memory centered on verified subtask-level state transitions.

The memory system consists of four complementary stores:
\begin{itemize}
    \item \textbf{Working memory} records the current objective, active subtask, verified progress, unresolved requirements, and recent failure or no-progress states.
    
    \item \textbf{Episodic memory} records the state transition associated with each bounded VLA invocation, together with the selected subtask, harness-verified outcome, execution cost, and diagnostic information.
    
    \item \textbf{Procedural memory} stores reusable dependencies among parameterized subtasks extracted from verified successful executions.
    
    \item \textbf{World memory} stores visually inferred landmarks and spatial relations that may support later planning within the current world.
\end{itemize}

The episodic record generated by bounded invocation $j$ is represented as
\begin{equation}
m_j
=
\left(
h_j,
u_j,
v_j,
h_{j+1},
c_j
\right),
\label{eq:appendix_episodic_memory}
\end{equation}
where $c_j$ contains execution metadata such as execution cost and diagnostic information.

At each high-level decision step, the current working memory $h_j$ is directly included in the planning context. Task-relevant entries from the other memory stores are retrieved according to
\begin{equation}
R_j
=
\operatorname{Retrieve}
\left(
G,
h_j,
\mathcal{M}_{\mathrm{epi}},
\mathcal{M}_{\mathrm{proc}},
\mathcal{M}_{\mathrm{world}}
\right).
\label{eq:appendix_memory_retrieval}
\end{equation}

Only visually verified outcomes are committed as successful task progress. Procedural memory is updated only from verified successful executions.

To reduce memorization of specific evaluation trajectories, procedural memory does not retain absolute coordinates or fixed camera orientations. Memory accumulated during the current episode remains available for online planning, whereas episodic records from previous runs of the same evaluation seed are excluded from retrieval. This prevents direct replay of routes discovered during earlier evaluations of the same seed.

\subsection{Matched Harness Comparison}
\label{app:harness_comparison}

To isolate the contribution of agentic orchestration from lower-level policy quality, we compare harness variants while holding the lower-level VLA and upper-level semantic model fixed.

The fixed-horizon baseline executes the same atomic instruction for a predetermined control horizon before returning control to the upper level. Full HiSTEVE instead uses state-dependent handoff, visual verification, structured memory update, local recovery, and replanning.

Let $\pi_\theta$ denote the shared VLA policy and $\mathcal{H}$ the high-level harness. The comparison can therefore be written as
\begin{equation}
\pi_{\mathrm{agent}}
=
\mathcal{H}
\left(
\pi_\theta
\right),
\label{eq:appendix_agent_composition}
\end{equation}
where $\pi_\theta$ is held fixed and only $\mathcal{H}$ changes between the compared systems.

This controlled comparison produces only a small change in atomic-task performance: MineStudio All changes from $81.2\%$ to $85.0\%$, while OpenHA Avg. changes from $37.8\%$ to $37.4\%$. In contrast, LongBench goal success increases from $10.7\%$ to $25.0\%$, and diagnostic progress increases from $17.9\%$ to $28.6\%$.

The contrast between nearly unchanged atomic performance and substantially improved long-horizon performance indicates that the primary gain arises from how the same controller is scheduled, monitored, verified, and recovered over extended interaction rather than from a stronger underlying VLA.

\section{VLA Architecture and Action Representation Details}
\label{app:vla_architecture}

\subsection{Cross-Regime Gradient Conflict}
\label{app:gradient_conflict}

This subsection provides the complete experimental setup and results for the cross-regime gradient-conflict analysis summarized in the main text. We train a fully shared baseline without regime-specific parameter branches, keeping the vision encoder frozen while jointly optimizing the remaining decoder parameters on the union of non-GUI and GUI data.

\paragraph{Gradient Conflict Gap Measurements.}
During the $30{,}000$-step training run, we establish a probe point every $200$ optimizer steps, including at initialization. At each probe point, we hold the current parameters fixed and compute gradients separately on pure non-GUI and pure GUI mini-batches without applying an optimizer update.
At each of the $28$ decoder layers, we extract the gradients of the four attention projections, the two RMSNorm modules, and the feed-forward network, and flatten the gradient of each component into a single vector.
Each gradient estimate is accumulated over $8$ mini-batches to reduce batch-level noise. Data augmentation is disabled during probing, and samples are drawn from a fixed probe split excluded from parameter optimization.

Let $g$ and $h$ denote the gradients of the same model component obtained from GUI and non-GUI data, respectively. Their cross-regime cosine similarity is
\begin{equation}
c_{\mathrm{cross}}
=
\cos(g,h)
=
\frac{g^\top h}
{\lVert g\rVert_2\lVert h\rVert_2}.
\label{eq:cross_regime_cosine}
\end{equation}
The cross-regime cosine alone does not distinguish regime differences from mini-batch variability. We therefore divide the probe data for each regime into two disjoint subsets, $A$ and $B$, and compute the corresponding within-regime similarities
\begin{equation}
c_{\mathrm{GUI}}
=
\cos(g_A,g_B),
\qquad
c_{\mathrm{NG}}
=
\cos(h_A,h_B).
\label{eq:within_regime_cosine}
\end{equation}
We define the conflict gap as the difference between the mean within-regime similarity and the cross-regime similarity:
\begin{equation}
\Delta
=
\frac{1}{2}
\left(
c_{\mathrm{GUI}}+c_{\mathrm{NG}}
\right)
-
c_{\mathrm{cross}}.
\label{eq:gradient_conflict_gap}
\end{equation}
A positive $\Delta$ therefore indicates that cross-regime gradients are less aligned than gradients from the same regime, relative to the within-regime reference induced by mini-batch variability.

\begin{figure*}[t]
    \centering
    \includegraphics[
        width=\textwidth
    ]{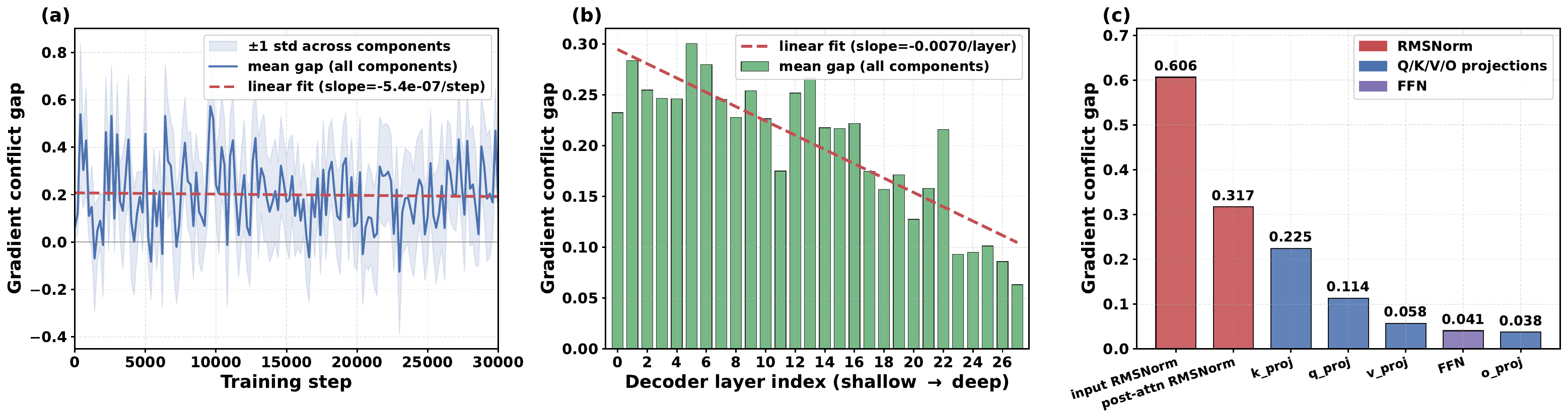}
    \caption{
        Overview of the cross-regime gradient-conflict analysis.
        (a) Mean conflict gap over all layers and components throughout training; the shaded region denotes $\pm1$ standard deviation across components, and the dashed line is a linear fit.
        (b) Mean conflict gap by decoder depth, averaged over training steps and components.
        (c) Mean conflict gap by shared component, averaged over training steps and decoder layers.
        All panels are computed from the same set of gradient probes.
    }
    \label{fig:gradient_conflict}
\end{figure*}

\paragraph{Conflict throughout Training.}
As shown in Figure~\ref{fig:gradient_conflict}(a), the conflict gap averaged over all layers and components has an overall mean of $0.1997$. The fitted slope is $-5.4\times10^{-7}$ per optimizer step, corresponding to a total change of approximately $-0.016$ over $30{,}000$ steps, indicating no substantial reduction in cross-regime misalignment during training.
At initialization, $\Delta$ is approximately $0.06$. The gap increases during the first several hundred optimizer steps and persists throughout the remainder of training, suggesting that it is not restricted to a short transient near initialization.

\paragraph{Conflict across Decoder Depth.}
Figure~\ref{fig:gradient_conflict}(b) shows that the conflict gap decreases with decoder depth when averaged over training steps and components, with a fitted slope of $-0.007$ per layer. The early layers L0--L8 have a mean gap of $0.257$, approximately twice the mean of $0.124$ in the deeper layers L19--L27, while the mean gap remains positive at every decoder layer.
This result shows that cross-regime misalignment is stronger in the early layers but is not confined to a small subset of the decoder. The pattern is consistent with the two regimes placing different demands on lower-level visuomotor features, although the analysis alone does not establish a causal source.

\paragraph{Conflict across Shared Components.}
As shown in Figure~\ref{fig:gradient_conflict}(c), the two RMSNorm modules exhibit the highest mean conflict gaps: $0.606$ for the input RMSNorm and $0.317$ for the post-attention RMSNorm. Among the remaining components, the K projection has a gap of $0.225$, while the Q, V, and O projections and the feed-forward network have gaps of $0.114$, $0.058$, $0.038$, and $0.041$, respectively.
Because the learnable parameters of RMSNorm are channel-wise scale vectors, the concentration of conflict in these modules suggests that regime-specific feature scaling provides a parameter-efficient location for specialization. This observation motivates the regime-specific RMSNorm branches introduced in the main method.

\subsection{Backbone and Regime Routing}
\label{app:backbone_routing}

The VLA policy is instantiated from Qwen2.5-VL-7B \citep{bai2025qwen25vltechnicalreport}, retaining its original vision encoder, multimodal projector, and autoregressive decoder. The current visual observation and atomic-subtask instruction are serialized using the native Qwen2.5-VL multimodal template. Action symbols are assigned to reserved vocabulary entries that do not overlap with natural-language tokens, allowing actions to be generated through the original autoregressive output head.

Each atomic subtask is associated with an interaction regime
$c_t\in\{\mathrm{NG},\mathrm{GUI}\}$. This regime serves as an explicit routing variable that selects the corresponding RMSNorm branch and mouse-displacement token set. The same regime is serialized as the interaction-type token after the beginning-of-action token. Apart from the regime-specific RMSNorm parameters and mouse-token embeddings, the remaining backbone parameters are shared across the two regimes.

\subsection{Action Representation}
\label{app:action_representation}

HiSTEVE represents native keyboard-and-mouse inputs as discrete action tokens, without using symbolic actions or arbitrary text input. Following the mouse encoding used in prior Minecraft policies \citep{baker2022video,li2025jarvis}, we apply $\mu$-law companding to the horizontal and vertical mouse displacements independently and quantize each transformed axis into $21$ bins.

Non-GUI and GUI interactions use disjoint mouse-token sets. Mouse displacement controls camera orientation in the non-GUI regime and cursor position in the GUI regime. The two regimes share the same numerical displacement bins, but each bin is assigned a different vocabulary entry with an independently learned embedding. Table~\ref{tab:action_space} summarizes the resulting native action space.

\begin{table*}[t]
    \centering
    \caption{
        Native Minecraft action space used by HiSTEVE. Non-GUI and GUI mouse movements share the same displacement codebook but are represented by disjoint token sets.
    }
    \label{tab:action_space}
    \footnotesize
    \setlength{\tabcolsep}{6pt}
    \renewcommand{\arraystretch}{1.02}
    \begin{tabularx}{\textwidth}{@{}c l l X@{}}
        \toprule
        \textbf{Index}
        & \textbf{Action}
        & \textbf{Human Input}
        & \textbf{Description} \\
        \midrule
        1  & Forward          & key \texttt{W}              & Move forward. \\
        2  & Back             & key \texttt{S}              & Move backward. \\
        3  & Left             & key \texttt{A}              & Strafe left. \\
        4  & Right            & key \texttt{D}              & Strafe right. \\
        5  & Jump             & key \texttt{Space}          & Jump or swim upward. \\
        6  & Sneak            & key \texttt{Left Shift}     & Sneak and move cautiously. \\
        7  & Sprint           & key \texttt{Left Ctrl}      & Sprint in the current movement direction. \\
        8  & Attack           & \texttt{Left Mouse Button}  & Attack an entity or break a block. \\
        9  & Use              & \texttt{Right Mouse Button} & Interact with an object. \\
        10 & Inventory        & key \texttt{E}              & Open or close the inventory. \\
        11 & Hotbar: $[1$--$9]$ & keys \texttt{1--9}        & Select the corresponding hotbar slot. \\
        12 & Non-GUI Mouse X  & move \texttt{Mouse X}       & Rotate the camera horizontally. \\
        13 & Non-GUI Mouse Y  & move \texttt{Mouse Y}       & Rotate the camera vertically. \\
        14 & GUI Mouse X      & move \texttt{Mouse X}       & Move the GUI cursor horizontally. \\
        15 & GUI Mouse Y      & move \texttt{Mouse Y}       & Move the GUI cursor vertically. \\
        \bottomrule
    \end{tabularx}
\end{table*}

The native controls in Table~\ref{tab:action_space} are serialized into a fixed-order action-token sequence. The beginning-of-action, end-of-action, and interaction-regime tokens define the sequence structure and do not introduce additional native input events. For each control step, the sequence is
\begin{equation}
\mathbf{a}_t
=
\Big[
\langle\mathrm{BOA}\rangle,\,
\langle c_t\rangle,\,
g_t^{(1)},\ldots,g_t^{(M)},\,
m_{x,b_{t,x}}^{(c_t)},\,
m_{y,b_{t,y}}^{(c_t)},\,
\langle\mathrm{EOA}\rangle
\Big],
\qquad
c_t\in\{\mathrm{NG},\mathrm{GUI}\},
\label{eq:action_serialization}
\end{equation}
where $g_t^{(1)},\ldots,g_t^{(M)}$ denote the discrete key and button groups in their fixed serialization order, and
$b_{t,x},b_{t,y}\in\{1,\ldots,21\}$ are the quantized horizontal and vertical mouse-displacement indices.

\section{VLA Training Strategy Details}
\label{app:vla_training}

\subsection{World Knowledge Injection}
\label{app:world_knowledge_details}

To reliably map subtasks to low-level actions, the policy model must acquire knowledge of Minecraft's entities, blocks and crafting mechanics. We therefore inject world knowledge by supervised fine-tuning.
We adopt the world knowledge and alignment data released with JARVIS-VLA \citep{li2025jarvis}, and deduplicate it to remove near-duplicate samples.
We freeze the vision encoder, and fully finetune the language decoder, so that it acquires game world knowledge and aligns it with first-person visual observations.

\paragraph{Training Sample Construction.}

The world knowledge data contain three supervision formats: text-only knowledge question answering, captioning of in-game screenshots, and visual question answering over the screenshots. Table~\ref{tab:world_knowledge_samples} illustrates the model input and supervised target.

\paragraph{Data Curation.}

Before training, we apply format-specific normalization, deduplication, and quality filtering; for knowledge-QA targets, we capitalize the initial letter and add missing sentence-final punctuation.
For exact deduplication, we lowercase and whitespace-normalize questions and retain the longest target for each normalized question.
For knowledge QA, we use FAISS \citep{johnson2019billion} to retrieve candidate neighbors from \texttt{all-MiniLM-L6-v2} sentence embeddings \citep{reimers2019sentence,wang2020minilm} and merge candidate pairs with cosine similarity of at least $0.95$ via union--find; for VQA, both text similarity from the same sentence encoder and image similarity computed from CLIP ViT-B/32 embeddings \citep{radford2021learning} must be at least $0.85$.
Finally, we remove QA samples with empty questions or targets, or targets containing at most five words excluding punctuation.

\paragraph{Training Objective.}

Let $x$ denote either a text-only context or an image--text context, and let $y=(y_1,\ldots,y_{|y|})$ denote the corresponding target answer or caption. World knowledge injection uses an autoregressive cross-entropy objective evaluated only over the target-output tokens:
\begin{equation}
\mathcal{L}_{\mathrm{WK}}(\theta)
=
-
\mathbb{E}_{(x,y)\sim\mathcal{D}_{\mathrm{WK}}}
\left[
\sum_{j=1}^{|y|}
\log
p_{\theta}
\left(
y_j
\mid
x,y_{<j}
\right)
\right],
\label{eq:world_knowledge_objective}
\end{equation}
where $\mathcal{D}_{\mathrm{WK}}$ denotes the mixed world knowledge dataset after filtering and deduplication. Input-prompt and visual tokens are excluded from the loss.

\subsection{Imitation Learning}
\label{app:imitation_details}

Starting from the world knowledge checkpoint, we train the policy using VPT gameplay data \citep{baker2022video}, trajectories released with JARVIS-VLA \citep{li2025jarvis}, and script-generated trajectories that supplement crafting and smelting recipes.
After deduplication, we resample trajectories by task to balance the task distribution.
Each training example consists of the subtask instruction, the single-frame observation, and the corresponding keyboard-and-mouse action.
At this stage, we duplicate each RMSNorm scale vector to initialize the GUI and non-GUI branches, freeze the vision encoder, and finetune all remaining policy parameters.

\paragraph{Training Sample Construction.}

Starting from the world knowledge checkpoint, we divide demonstration trajectories into atomic-task clips and construct each training example from a single-frame visual observation, a subtask instruction, an interaction regime, and a target action sequence. Table~\ref{tab:imitation_samples} shows the corresponding formats for non-GUI and GUI training examples.

\paragraph{Data Curation.}

To improve recipe coverage, we first add $10$ programmatically generated training trajectories for each \texttt{craft\_item} or \texttt{smelt\_item} recipe absent from the collected trajectories.
We then resample the augmented dataset at the atomic-task-clip level to obtain a $1{:}1$ GUI-to-non-GUI ratio, while ensuring that each designated major task family (e.g., \texttt{kill\_entity} and \texttt{smelt\_item}) accounts for at least $10\%$ of all training samples.

\paragraph{Training Objective.}

Let $o_t$ denote the current visual observation, $u_t$ the subtask instruction, $c_t\in\{\mathrm{NG},\mathrm{GUI}\}$ the interaction regime, and $\mathbf a_t=(a_{t,1},\ldots,a_{t,|\mathbf a_t|})$ the target action sequence. Imitation learning uses an autoregressive cross-entropy objective evaluated only over the target action tokens:
\begin{equation}
\mathcal{L}_{\mathrm{IL}}(\theta)
=
-
\mathbb{E}_{
(o_t,u_t,c_t,\mathbf a_t)
\sim
\mathcal{D}_{\mathrm{IL}}
}
\left[
\sum_{k=1}^{|\mathbf a_t|}
\log
p_{\theta}
\left(
a_{t,k}
\mid
\mathbf a_{t,<k},
o_t,
u_t,
c_t
\right)
\right],
\label{eq:imitation_objective}
\end{equation}
where $c_t$ selects the corresponding RMSNorm branch and mouse-displacement token set, while the interaction-type identifier token remains part of $\mathbf a_t$ and participates in the training loss.

\subsection{Reinforcement Learning}
\label{app:rl_details}

\paragraph{Rollout Construction.}
For each same-start branch group in Equation~\ref{eq:branch_group_advantage}, the initial condition fixes the task, world seed, initial inventory, and GUI configuration.
Each branch is executed in a separate Minecraft process to prevent cross-branch state contamination. We record visual observations, action tokens, behavior-policy log-probabilities, parsed environment actions, termination reasons, and trainer-side environment evidence.
For local comparisons, we prioritize candidate segments from the same branch group whose structured state signatures match at the beginning of the segment.

\paragraph{Environment-Based Rewards.}
Post-training determines atomic-subtask completion from environment state changes rather than model-reported success.
For an item-acquisition task, the completion certificate for target item $g$ is
\begin{equation}
C_t^g
=
\mathbf{1}
\left[
N_t(g)-N_0(g)
\geq
\delta_g
\right],
\label{eq:terminal_certificate}
\end{equation}
where $N_t(g)$ is the inventory count at step $t$ and $\delta_g$ is the required increase relative to the initial state.
Thus, breaking a target block without collecting the drop, or producing an item in a GUI without transferring it to the player inventory, is not considered successful.

For GUI tasks, required-material coverage, correct slot occupancy, process initiation, output appearance, incorrect products, and material loss are summarized by a progress potential $\Phi(s)$.
The step reward combines the change in progress potential with a terminal reward issued when the completion certificate is first satisfied:
\begin{equation}
r_t
=
\Phi(s_{t+1})
-
\Phi(s_t)
+
R_T
\mathbf{1}
\left[
C_{t+1}^g
\land
\neg C_t^g
\right],
\label{eq:potential_reward}
\end{equation}
where $R_T$ is the terminal success reward. Taking differences in $\Phi$ prevents the policy from repeatedly receiving the same progress reward by remaining at an already reached stage.

\paragraph{Branch Optimization.}
Let $x_i$ denote the policy-visible visual--language condition associated with sample $i$, and let $\mathbf{a}_i$ denote the optimized action sequence. The behavior-policy correction weight and current-policy update ratio are
\begin{equation}
w_i
=
\frac{
\pi_{\mathrm{old}}(\mathbf{a}_i\mid x_i)
}{
\pi_b(\mathbf{a}_i\mid x_i)
},
\qquad
\rho_i
=
\frac{
\pi_\theta(\mathbf{a}_i\mid x_i)
}{
\pi_{\mathrm{old}}(\mathbf{a}_i\mid x_i)
}.
\label{eq:policy_ratios}
\end{equation}
We detach $w_i$ from gradient computation and use $\rho_i$ for the clipped group-relative update:
\begin{equation}
\mathcal{L}_{\mathrm{GRPO}}
=
-
\mathbb{E}_i
\left[
w_i
\min
\left(
\rho_i A_i,
\operatorname{clip}
\left(
\rho_i,
1-\varepsilon_{\mathrm{clip}},
1+\varepsilon_{\mathrm{clip}}
\right)
A_i
\right)
\right].
\label{eq:branch_grpo}
\end{equation}

For locally matched segments, let $(x^+,\mathbf{a}^+)$ and $(x^-,\mathbf{a}^-)$ denote the preferred and rejected segments from successful and failed branches, respectively.
The corresponding pairwise objective is
\begin{equation}
\mathcal{L}_{\mathrm{pair}}
=
-
\log
\sigma
\left(
\beta_{\mathrm{pair}}
\left[
\log
\frac{
\pi_\theta(\mathbf{a}^{+}\mid x^{+})
}{
\pi_{\mathrm{ref}}(\mathbf{a}^{+}\mid x^{+})
}
-
\log
\frac{
\pi_\theta(\mathbf{a}^{-}\mid x^{-})
}{
\pi_{\mathrm{ref}}(\mathbf{a}^{-}\mid x^{-})
}
\right]
\right).
\label{eq:state_pairwise}
\end{equation}
Equations~\ref{eq:branch_grpo} and~\ref{eq:state_pairwise} instantiate $\mathcal{L}_{\mathrm{branch}}$ in Equation~\ref{eq:post_training_objective} according to the available comparison.

\section{Task Completion State Detector Details}
\label{app:detector_details}

\subsection{Dataset Construction and Evaluation Split}
\label{app:detector_dataset}

We construct the completion detection dataset from VPT human gameplay trajectories, using offline event and inventory-state detection to identify the task category and completion time of each atomic-task segment. Source trajectories are partitioned before temporal observation windows are sampled, ensuring that no windows from the same trajectory appear across data splits.

For each atomic-task segment, let $t^\star$ denote the detected completion time. A temporal sample queried at time $t$ receives the binary label
\begin{equation}
y_t
=
\mathbf{1}
\left[
t \ge t^\star
\right].
\label{eq:appendix_detector_label}
\end{equation}

Samples before $t^\star$ are labeled incomplete, while samples at or after $t^\star$ are labeled complete. Incomplete samples are downsampled to balance the two classes.
The reported validation set contains $4{,}674$ samples spanning mining, killing, crafting, and smelting.
All compared models receive the same temporal observations and subtask instruction.

\paragraph{Cross-Attention Implementation.}
Cross-attention modules are inserted after detector decoder layers
$\ell\in\{4,8,12,16,20,24,28,32,36\}$.
Expanding the multi-head attention in Equation~\ref{eq:detector-cross-attention}, the output of attention head $r$ and the resulting residual update are
\begin{equation}
\begin{aligned}
U_{\ell,r}
&=
\operatorname{softmax}
\left(
\frac{
\left(\operatorname{LN}_{\ell}(Z_t^{(\ell)})W_{\ell,r}^{Q}\right)
\left(H_t^\pi W_{\ell,r}^{K}\right)^{\top}
}{
\sqrt{d_h}
}
\right)
\left(H_t^\pi W_{\ell,r}^{V}\right),
\\
\widetilde{Z}_t^{(\ell)}
&=
Z_t^{(\ell)}
+
\operatorname{Concat}_{r=1}^{n_h}
\left(U_{\ell,r}\right)
W_{\ell}^{O},
\end{aligned}
\label{eq:detector-cross-attention-implementation}
\end{equation}
where $n_h$ and $d_h$ denote the number of attention heads and the dimension of each head, respectively.

\paragraph{Channel Pruning.}
We estimate feed-forward channel importance using intermediate activations collected from calibration samples and the corresponding down-projection weights.
Let $A_\ell\in\mathbb{R}^{N\times d_{\mathrm{ff}}}$ denote the intermediate activations at layer $\ell$, and let $W_\ell^{\mathrm{down}}\in\mathbb{R}^{d\times d_{\mathrm{ff}}}$ denote its down-projection matrix.
The channel score and retained set are defined as
\begin{equation}
s_{\ell,j}
=
\left\|A_{\ell,:,j}\right\|_2
\left\|W_{\ell,:,j}^{\mathrm{down}}\right\|_2,
\qquad
\mathcal{I}_\ell
=
\operatorname{TopK}
\left(
\{s_{\ell,j}\}_{j=1}^{d_{\mathrm{ff}}},
d'_{\mathrm{ff}}
\right),
\quad
d'_{\mathrm{ff}}
\approx
\frac{d_{\mathrm{ff}}}{3}.
\label{eq:detector-channel-pruning}
\end{equation}
The same channel indices are retained in the output dimensions of the gate and up projections and the corresponding input dimensions of the down projection.

\paragraph{Training Objective.}
During training, we freeze the VLA and stop gradients through $H_t^\pi$. We optimize only the cross-attention modules, classification head, and LoRA adapters applied to the detector's self-attention Q/K/V/O projections.
Let $y_t\in\{0,1\}$ denote the completion label and $p_{t,c}$ the predicted probability of class $c$. The detector is trained using label-smoothed two-class cross-entropy:
\begin{equation}
\mathcal{L}_{\mathrm{det}}
=
-\frac{1}{|\mathcal{B}|}
\sum_{t\in\mathcal{B}}
\sum_{c\in\{0,1\}}
\left[
(1-\epsilon)\mathbf{1}\{c=y_t\}
+\frac{\epsilon}{2}
\right]
\log p_{t,c},
\label{eq:detector-objective}
\end{equation}
where $\epsilon$ is the label-smoothing coefficient and $\mathcal{B}$ denotes a training batch.

\subsection{Deployment Decision Rule}
\label{app:detector_deployment}

The detector is evaluated before the VLA decodes the next action. Let $p_t$ denote the predicted probability of the completion class.
A single positive prediction does not immediately terminate the current invocation. Instead, the detector emits a candidate completion signal only if
\begin{equation}
p_{t-1} > \tau
\quad \land \quad
p_t > \tau.
\label{eq:appendix_detector_two_step}
\end{equation}

The two-query confirmation suppresses isolated false-positive predictions. If the condition is not satisfied, the VLA proceeds to decode and execute the next native action.

A candidate completion signal only returns control to the harness. It is not treated as direct task-success evidence. The harness subsequently verifies the corresponding visual completion conditions before updating the planning state.

\section{Experimental Protocol Details}
\label{app:evaluation_protocol}

\subsection{Atomic-Task Evaluation}
\label{app:atomic_evaluation}

We evaluates the lower-level VLA independently, while the harness ablation in Table~\ref{tab:main_ablation} evaluates system-level execution on the same atomic tasks.

\paragraph{MineStudio Bench.}
MineStudio Bench contains $20$ atomic tasks, evenly divided among Mine, Kill, Craft, and Smelt. Each task is evaluated with three independent trials. We report strict pass@1, which measures whether the policy succeeds on its first execution attempt. Mine and Kill primarily evaluate first-person world interaction, whereas Craft and Smelt additionally require GUI-mediated manipulation.

For category scores $S_{\mathrm{Mine}}$, $S_{\mathrm{Kill}}$, $S_{\mathrm{Craft}}$, and $S_{\mathrm{Smelt}}$, the overall score is
\begin{equation}
S_{\mathrm{MineStudio}}
=
\frac{
S_{\mathrm{Mine}}
+
S_{\mathrm{Kill}}
+
S_{\mathrm{Craft}}
+
S_{\mathrm{Smelt}}
}{4}.
\label{eq:minestudio_average}
\end{equation}

\subsection{LongBench Benchmark and Evaluation}
\label{app:longbench}
\label{app:long_horizon_evaluation}

LongBench evaluates whether atomic native-control capabilities can be composed into persistent behavior over extended interaction. In contrast to atomic benchmarks, each LongBench episode provides only a final goal: the agent must determine the required intermediate objectives online from its visual observations and execution history. Neither a reference task decomposition nor the benchmark's diagnostic annotations are exposed to the agent.

\paragraph{Task suite.}
LongBench contains $14$ goal-only tasks organized into four families that emphasize different long-horizon requirements: Resource \& Crafting, Exploration \& Memory, Recovery, and Open-ended interaction. Table~\ref{tab:longbench_tasks} summarizes the task suite.

\begin{table*}[t]
\centering
\small
\setlength{\tabcolsep}{5.0pt}
\renewcommand{\arraystretch}{1.08}
\caption{
LongBench task suite. Each episode exposes only the final goal to the agent; the decomposition and diagnostic checkpoints are withheld during execution.
}
\label{tab:longbench_tasks}
\begin{tabular}{lll}
\toprule
\textbf{Family} & \textbf{Task} & \textbf{Primary long-horizon requirement} \\
\midrule

\multirow{4}{*}{Resource \& Crafting}
& \texttt{wooden\_age}
& Resource acquisition and tool progression \\
& \texttt{stone\_age}
& Multi-stage gathering and crafting \\
& \texttt{iron\_tool}
& Resource, furnace, and tool-production chain \\
& \texttt{glass\_maker}
& Gathering, smelting, and output retrieval \\

\midrule

\multirow{4}{*}{Exploration \& Memory}
& \texttt{cave\_survey}
& Exploration and persistent environmental state \\
& \texttt{camp\_return}
& Landmark memory and return navigation \\
& \texttt{field\_camp}
& Exploration with stateful task continuation \\
& \texttt{resource\_revisit}
& Remembering and revisiting a prior location \\

\midrule

\multirow{3}{*}{Recovery}
& \texttt{tool\_break\_recovery}
& Recovery after tool failure \\
& \texttt{lost\_drop\_recovery}
& Recovery from lost task resources \\
& \texttt{fuel\_alternative}
& Substitution after unavailable intended resource \\

\midrule

\multirow{3}{*}{Open-ended}
& \texttt{cook\_dinner}
& Flexible multi-stage goal completion \\
& \texttt{light\_up}
& Open-ended resource acquisition and execution \\
& \texttt{survival\_iron\_ingot}
& Extended survival and resource progression \\

\bottomrule
\end{tabular}
\end{table*}

The four families serve complementary purposes. Resource \& Crafting tasks require the agent to compose multiple native-control skills into a dependency chain whose final object is unavailable at initialization. Exploration \& Memory tasks require information gathered earlier in the episode to affect later decisions. Recovery tasks introduce situations in which the current execution path can no longer be continued directly and therefore require local recovery or replanning. Open-ended tasks admit multiple valid strategies and provide less structure about the sequence of intermediate objectives.

\paragraph{Benchmark initialization.}

Eleven of the fourteen tasks start from an empty inventory in a natural survival world. Formal evaluation uses fixed environment seeds, while terrain, resource distribution, and naturally occurring entities remain part of the Minecraft environment.

The three Recovery tasks use controlled initial states to reproducibly instantiate low-frequency recovery scenarios involving tool breakage, alternative fuel, and missed resource pickup.

These states are instantiated only during reset; subsequent evaluation follows the same native visual and keyboard-and-mouse interface used throughout the paper.

\paragraph{Hidden diagnostic nodes.}
LongBench additionally defines $49$ hidden diagnostic nodes across the $14$ tasks. A diagnostic node denotes an interpretable intermediate state, such as obtaining a prerequisite resource, constructing an intermediate tool, reaching a relevant landmark, or entering a required processing stage.

These nodes are never included in the planner prompt and are not used to select the next subtask. They are recorded only by the evaluation procedure after execution and therefore provide a diagnostic view of trajectory progress without supplying privileged guidance to the agent.

For task $i$, let $\mathcal{D}_i$ denote its diagnostic-node set and $\mathcal{R}_i \subseteq \mathcal{D}_i$ the nodes reached during execution. We define diagnostic progress as

\begin{equation}
P_i
=
\frac{
|\mathcal{R}_i|
}{
|\mathcal{D}_i|
}.
\label{eq:longbench_task_progress}
\end{equation}

Across an evaluation set of $N$ instances, the aggregate progress metric is

\begin{equation}
\operatorname{Progress}
=
\frac{
\sum_{i=1}^{N}
|\mathcal{R}_i|
}{
\sum_{i=1}^{N}
|\mathcal{D}_i|
}.
\label{eq:longbench_progress}
\end{equation}

Diagnostic progress is not a success criterion. A trajectory may reach most intermediate diagnostic states but is still counted as unsuccessful if the final-goal predicate is not satisfied.

\paragraph{Evaluation interface.}

LongBench follows the same human-interface execution constraint used throughout HiSTEVE. At the beginning of each episode, the agent receives a natural-language final goal, while the deployed planner and VLA operate from first-person visual observations and interact through native keyboard-and-mouse actions.
Reference decompositions, hidden diagnostic nodes, and privileged environment state are not exposed to the planner or VLA.

\paragraph{Final-goal evaluation.}
The primary LongBench metric is final-goal success. For each task family, Goal SR is computed over its formal evaluation episodes. For evaluation instance $i$, let $g_i\in\{0,1\}$ denote whether the terminal task predicate is satisfied. We report

\begin{equation}
\operatorname{GoalSR}
=
\frac{1}{N}
\sum_{i=1}^{N}
g_i.
\label{eq:longbench_goal_success}
\end{equation}

Table~\ref{tab:main_results} reports the family-level Goal SR, while the overall Goal SR in Table~\ref{tab:main_ablation} is the macro-average over the four task families.

Final-goal success is authoritative: reaching intermediate milestones or producing partial progress does not convert an unsuccessful trajectory into a successful one.
In addition to Goal SR and diagnostic progress, we record environment steps, upper-level planner calls, detector-triggered handoffs, local recovery attempts, and replanning events. These quantities are used for trajectory analysis rather than as optimization objectives.

\paragraph{Matched harness evaluation.}
To isolate the effect of agent orchestration from lower-level policy quality, the harness ablation keeps the native-control VLA and upper-level semantic model fixed. The compared systems therefore share the same local controller and differ only in how repeated VLA invocations are organized over time.

The fixed-horizon variant returns control after a predetermined execution boundary. Full HiSTEVE instead uses state-dependent handoff followed by visual verification and planning-state update; incomplete executions may trigger local recovery or replanning.
Because the lower-level VLA is unchanged, this comparison isolates the effect of runtime orchestration from changes in policy weights.
The corresponding controlled results are reported in Table~\ref{tab:main_ablation}.

\paragraph{Diagnostic failure patterns.}
LongBench exposes failures that are less visible in atomic benchmarks because successful completion depends on preserving usable state over many policy invocations. We observe three recurring patterns.

First, multi-step GUI interactions can leave the environment in a partially correct state that is difficult to continue from reliably. For example, an invocation may correctly place part of a recipe but fail to retrieve the output or recover from an incorrect slot configuration. A subsequent invocation then starts from a state that differs from the demonstrations seen by the policy.

Second, exploration tasks amplify errors in spatial memory. The agent may successfully discover a relevant area but later fail to recover sufficiently precise visual or spatial context for a return objective. These cases distinguish reaching a useful location once from maintaining information that remains actionable several planning steps later.

Third, recovery itself consumes execution budget. A local failure may be recoverable, yet repeated attempts can leave insufficient budget for the remaining dependency chain. LongBench therefore measures not only whether recovery is possible, but whether recovery is efficient enough to preserve the feasibility of the overall objective.

These observations explain why diagnostic progress can substantially exceed final-goal success: HiSTEVE often completes meaningful portions of a long trajectory before a later execution or recovery failure prevents satisfaction of the terminal predicate.

\subsection{Controlled Harness Evaluation}
\label{app:controlled_harness_evaluation}

The harness ablation is designed to separate improvements in agent orchestration from changes in lower-level policy quality. Across compared variants, we hold the native-control VLA, upper-level semantic model, task definitions, and evaluation protocol fixed.

The fixed-horizon variant executes the same atomic instruction until a predetermined execution boundary before returning control to the upper level. Full HiSTEVE instead uses state-dependent completion signals followed by visual verification, structured task-state update, local recovery, and replanning.

Because the underlying VLA is fixed, differences in long-horizon performance primarily reflect how the same local controller is organized across repeated invocations. This is also why we report atomic metrics for the same comparison: large changes in long-horizon behavior accompanied by only small changes in atomic performance provide evidence that the gain arises from orchestration rather than from a stronger controller.

\section{Ablation of the State Detector Design}
\label{app:detector_ablation}

We study three design choices in the state detector: FFN pruning, cross-attention density, and policy-context injection.
Our final configuration uses Wanda-inspired scores \citep{sun2024simple} to retain approximately one-third of each FFN, inserts cross-attention after every fourth detector layer, and supplies the same final-layer policy context $H_t^\pi$ to all cross-attention modules.
All variants use the Qwen2.5-VL-3B backbone and are evaluated on the same class-balanced validation set of $4{,}674$ samples.
Table~\ref{tab:detector_ablation} summarizes these controlled comparisons; each ablation variant changes only one factor, with all remaining settings fixed to the final configuration.

\begin{table}[!t]
\centering
\small
\setlength{\tabcolsep}{5.5pt}
\renewcommand{\arraystretch}{1.10}
\caption{
Ablation of the state detector design. Each group changes one design factor while holding the remaining settings fixed.
}
\label{tab:detector_ablation}
\begin{tabular}{l cccccc}
\toprule
Variant
& Acc $\uparrow$
& Prec $\uparrow$
& Rec $\uparrow$
& FPR $\downarrow$
& FNR $\downarrow$
& F1 $\uparrow$ \\
\midrule
\textbf{Ours}
& 0.863 & 0.861 & 0.865 & 0.140 & 0.135 & 0.863 \\
\midrule

\multicolumn{7}{l}{\emph{FFN intermediate width}} \\
\quad Wanda, width $\tfrac{1}{2}$
& 0.865 & 0.863 & 0.867 & 0.138 & 0.133 & 0.865 \\
\quad Wanda, width $\tfrac{1}{4}$
& 0.739 & 0.739 & 0.740 & 0.262 & 0.260 & 0.739 \\
\midrule

\multicolumn{7}{l}{\emph{FFN channel selection}} \\
\quad Prefix pruning, width $\tfrac{1}{3}$
& 0.768 & 0.768 & 0.768 & 0.232 & 0.232 & 0.768 \\
\midrule

\multicolumn{7}{l}{\emph{Cross-attention density}} \\
\quad Every $2$ layers
& 0.869 & 0.867 & 0.873 & 0.134 & 0.127 & 0.870 \\
\quad Every layer
& 0.859 & 0.859 & 0.859 & 0.141 & 0.141 & 0.859 \\
\midrule

\multicolumn{7}{l}{\emph{Policy-context injection}} \\
\quad Layer-wise pairing
& 0.690 & 0.690 & 0.690 & 0.310 & 0.310 & 0.690 \\
\bottomrule
\end{tabular}
\end{table}

\paragraph{Effect of FFN Width.}
Reducing the FFN width from one-half to one-third changes accuracy only from $0.865$ to $0.863$, whereas reducing it further to one-quarter decreases accuracy to $0.739$.
We therefore use one-third as the most compact tested width that preserves detection performance.

\paragraph{Effect of Channel Selection.}
At the same one-third width, naive prefix truncation reduces accuracy from $0.863$ to $0.768$.
This result demonstrates the importance of selecting retained channels according to their activation--weight importance.
We therefore use Wanda-inspired activation--weight scores for channel selection in the final detector.

\paragraph{Effect of Cross-Attention Density.}
Inserting cross-attention every two layers achieves an accuracy of $0.869$, whereas inserting it after every layer yields $0.859$, showing that greater density does not provide gains.
The every-fourth-layer configuration uses only $9$ modules, compared with $18$ and $36$ in the denser variants, while remaining within $0.006$ accuracy of the best result.
We therefore adopt the every-fourth-layer placement to balance detection performance and overhead.

\paragraph{Final-Layer Broadcast versus Layer-Wise Pairing.}
Using depth-matched intermediate policy representations reduces accuracy from $0.863$ to $0.690$.
This substantial gap supports broadcasting the final-layer policy context to all cross-attention modules, as it integrates information from the current observation and subtask.
We therefore broadcast the same final-layer policy context $H_t^\pi$ to all cross-attention modules in the final detector.

\section{Evaluation Variability}
\label{app:evaluation_variability}

Table~\ref{tab:main_results_variability} reports the mean and standard deviation of the benchmark results presented in Table~\ref{tab:main_results}.

\begin{table*}[!t]
\centering
\small
\setlength{\tabcolsep}{3pt}
\renewcommand{\arraystretch}{1.15}
\caption{Benchmark performance with variability on MineStudio Bench, OpenHA, and LongBench. Values are mean $\pm$ standard deviation in percentage points; bold and underlined means denote the best and second-best results within each column.}
\label{tab:main_results_variability}
\newcommand{\ms}[2]{\ensuremath{#1^{\pm #2}}}
\resizebox{\textwidth}{!}{%
\begin{tabular}{l*{12}{c}}
\toprule
 & \multicolumn{4}{c}{\textbf{MineStudio Bench}}
 & \multicolumn{4}{c}{\textbf{OpenHA}}
 & \multicolumn{4}{c}{\textbf{LongBench}} \\
\cmidrule(lr){2-5} \cmidrule(lr){6-9} \cmidrule(lr){10-13}
\textbf{Method}
& \textbf{Mine} & \textbf{Kill} & \textbf{Craft} & \textbf{Smelt}
& \textbf{Emb.} & \textbf{GUI} & \textbf{Combat} & \textbf{Avg.}
& \textbf{R\&C} & \textbf{E\&M} & \textbf{Rec.} & \textbf{Open} \\
\midrule
VPT
& \ms{20.0}{20.0} & \ms{13.3}{11.5} & \ms{0.0}{0.0} & \ms{6.7}{11.5}
& \ms{10.1}{4.1} & \ms{0.7}{1.1} & \ms{3.6}{1.5} & \ms{4.8}{1.5}
& \ms{0.0}{0.0} & \ms{1.4}{2.4} & \ms{0.0}{0.0} & \ms{0.0}{0.0} \\

STEVE-1
& \ms{20.0}{20.0} & \ms{13.3}{11.5} & \ms{0.0}{0.0} & \ms{6.7}{11.5}
& \ms{8.4}{2.5} & \ms{0.0}{0.0} & \ms{4.9}{2.6} & \ms{4.4}{1.5}
& \ms{0.0}{0.0} & \ms{0.7}{1.9} & \ms{0.0}{0.0} & \ms{0.0}{0.0} \\

ROCKET-1
& \ms{33.3}{11.5} & \ms{40.0}{20.0} & \ms{0.0}{0.0} & \ms{6.7}{11.5}
& \ms{19.2}{2.9} & \ms{0.0}{0.0} & \ms{29.8}{3.9} & \ms{16.3}{1.0}
& \ms{1.4}{3.8} & \ms{2.9}{3.9} & \ms{0.7}{1.9} & \ms{0.0}{0.0} \\

JARVIS-VLA
& \ms{66.7}{11.5} & \ms{60.0}{20.0} & \ms{26.7}{11.5} & \ms{60.0}{34.6}
& \ms{31.0}{4.6} & \ms{25.3}{3.6} & \ms{18.3}{4.3} & \ms{24.9}{1.9}
& \ms{7.1}{4.9} & \ms{5.7}{4.5} & \ms{1.4}{3.8} & \ms{0.7}{1.9} \\

LatentHA
& \ms{46.7}{23.1} & \ms{33.3}{30.6} & \ms{6.7}{11.5} & \ms{20.0}{20.0}
& \ms{27.3}{4.2} & \ms{3.5}{1.4} & \ms{8.2}{3.6} & \ms{13.0}{1.7}
& \ms{2.1}{3.9} & \ms{2.9}{2.7} & \ms{0.7}{1.9} & \ms{0.0}{0.0} \\

MotionHA
& \ms{53.3}{30.6} & \ms{33.3}{11.5} & \ms{6.7}{11.5} & \ms{20.0}{20.0}
& \ms{31.6}{3.2} & \ms{0.0}{0.0} & \ms{9.1}{2.1} & \ms{13.6}{1.4}
& \ms{2.9}{3.9} & \ms{3.6}{3.8} & \ms{0.7}{1.9} & \ms{0.0}{0.0} \\

GroundingHA
& \ms{60.0}{0.0} & \ms{53.3}{30.6} & \ms{13.3}{23.1} & \ms{33.3}{23.1}
& \ms{\underline{39.7}}{3.3} & \ms{3.7}{1.8} & \ms{28.2}{4.0} & \ms{23.9}{1.8}
& \ms{5.7}{7.3} & \ms{4.3}{3.5} & \ms{1.4}{2.4} & \ms{0.0}{0.0} \\

TextVLA
& \ms{46.7}{23.1} & \ms{46.7}{23.1} & \ms{13.3}{11.5} & \ms{40.0}{20.0}
& \ms{23.9}{5.3} & \ms{14.0}{3.9} & \ms{27.1}{5.2} & \ms{21.7}{3.0}
& \ms{7.1}{6.4} & \ms{5.0}{5.8} & \ms{1.4}{2.4} & \ms{0.7}{1.9} \\

OpenHA Text/CoA
& \ms{\underline{73.3}}{30.6} & \ms{\underline{73.3}}{11.5} & \ms{\underline{40.0}}{20.0} & \ms{\underline{73.3}}{30.6}
& \ms{37.0}{3.5} & \ms{\mathbf{33.3}}{3.9} & \ms{\mathbf{40.0}}{3.8} & \ms{\underline{36.8}}{2.0}
& \ms{\underline{14.3}}{7.9} & \ms{\underline{7.1}}{7.0} & \ms{\underline{2.9}}{3.9} & \ms{\underline{1.4}}{2.4} \\

\midrule
\textbf{HiSTEVE (Ours)}
& \ms{\mathbf{93.3}}{11.5} & \ms{\mathbf{86.7}}{11.5} & \ms{\mathbf{66.7}}{11.5} & \ms{\mathbf{93.3}}{11.5}
& \ms{\mathbf{46.0}}{2.9} & \ms{\underline{27.2}}{4.3} & \ms{\underline{39.0}}{3.7} & \ms{\mathbf{37.4}}{1.8}
& \ms{\mathbf{42.9}}{14.4} & \ms{\mathbf{28.6}}{12.5} & \ms{\mathbf{14.3}}{5.3} & \ms{\mathbf{14.3}}{10.2} \\
\bottomrule
\end{tabular}%
}
\end{table*}

\section{Discussion}

HiSTEVE suggests that a central challenge in long-horizon agency is maintaining task state across planning and control decisions. The agent harness provides this continuity by preserving task state, aligning plans with executable capabilities, and using execution outcomes to make decisions.

With the lower-level VLA fixed, the complete harness changes aggregate atomic performance only modestly while substantially improving LongBench goal success. This suggests that long-horizon capability depends not only on the strength of the local controller, but also on how existing capabilities are organized across repeated invocations, with task state preserved and subsequent decisions adapted to execution outcomes.
A further design insight is to decouple control handoff from task-success verification. The completion detector provides only a candidate handoff signal, while the harness subsequently verifies completion from visual evidence, preventing individual detector errors from being directly committed to persistent task state.

The current system has two main limitations. First, its atomic subtask interfaces are manually defined, and its capabilities remain limited by the existing VLA policy; adding new atomic capabilities requires additional training data and policy updates, while neither the harness nor the VLA currently performs parametric continual learning from deployment experience. In addition, high-level planning relies on an off-the-shelf VLM, making overall system performance dependent on the capabilities of the selected VLM backend.

Future work could enable independent or joint evolution of the VLA and harness: the VLA could learn continually from failure and recovery trajectories, while the harness updates its planning constraints and recovery strategies from experience. A complementary direction is to use Minecraft as training an environment for agent intelligence, examining whether planning, memory, and recovery capabilities can transfer across games and ultimately to interactive domains beyond games.

\begin{table*}[t]
    \centering
    \caption{
        Representative examples of the three data formats used for world knowledge injection.
    }
    \label{tab:world_knowledge_samples}

    \setlength{\fboxsep}{8pt}
    \fbox{%
    \begin{minipage}{0.955\textwidth}
        \small

        \noindent\textbf{Text QA Example}
        \par\vspace{3pt}
        \hrule
        \vspace{6pt}

        \noindent\textbf{Question:}
        How do you use enchantments to enhance items and equipment in Minecraft?

        \vspace{5pt}

        \noindent\textbf{Target answer:}
        By placing the item in an Enchantment Table to apply random enchantments, or by using an Anvil to combine items or apply enchanted books.

        \vspace{7pt}
        \hrule
        \vspace{7pt}

        \noindent\textbf{Caption Example}
        \par\vspace{3pt}
        \hrule
        \vspace{5pt}

        \begin{center}
            \includegraphics[
                width=0.5\linewidth
            ]{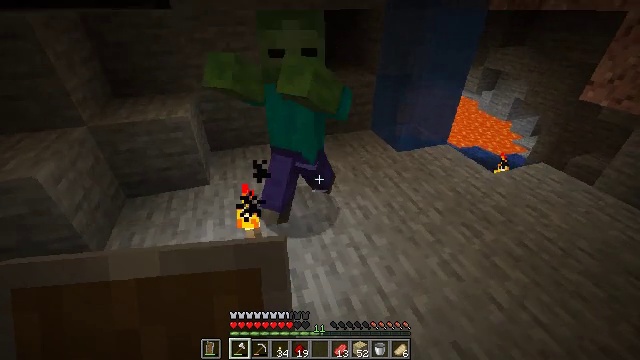}
        \end{center}

        \noindent\textbf{Instruction:}
        Please provide a caption for this Minecraft screenshot.

        \vspace{5pt}

        \noindent\textbf{Target caption:}
        A zombie stands centrally in front of the player in a cave. Stone walls surround the player and the zombie. A pool of lava is visible in the background of the cave.

        \vspace{7pt}
        \hrule
        \vspace{7pt}

        \noindent\textbf{Visual Question Answering Example}
        \par\vspace{3pt}
        \hrule
        \vspace{5pt}

        \begin{center}
            \includegraphics[
                width=0.5\linewidth
            ]{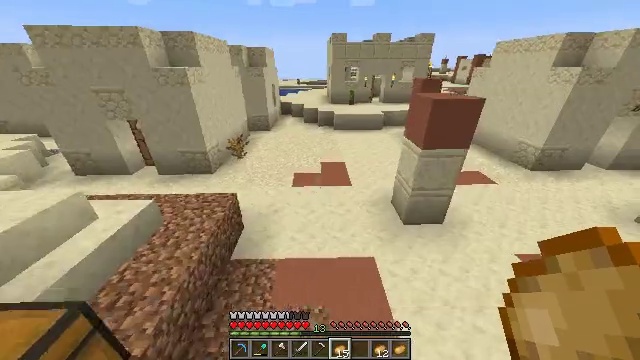}
        \end{center}

        \noindent\textbf{Question:}
        What type of biome is visible in the image?

        \vspace{5pt}

        \noindent\textbf{Target answer:}
        The biome visible in the image is a Desert biome.

    \end{minipage}%
    }
\end{table*}

\begin{table*}[t]
    \centering
    \caption{
        Representative imitation-learning examples from the non-GUI and GUI interaction regimes. Each example contains a single-frame observation, an atomic-subtask instruction, and its aligned target action sequence.
    }
    \label{tab:imitation_samples}

    \setlength{\fboxsep}{8pt}
    \fbox{%
    \begin{minipage}{0.955\textwidth}
        \small

        \noindent\textbf{Non-GUI Interaction Example}
        \par\vspace{3pt}
        \hrule
        \vspace{5pt}

        \begin{center}
            \includegraphics[
                width=0.5\linewidth
            ]{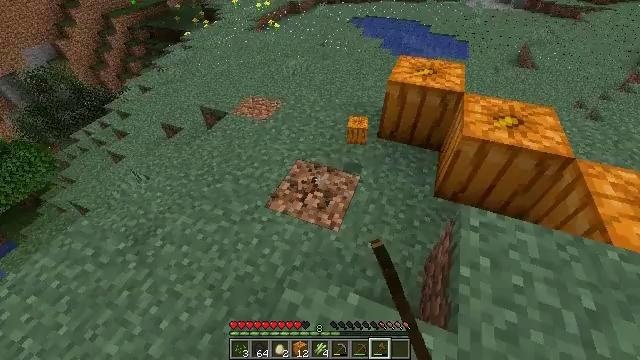}
        \end{center}

        \noindent\textbf{Subtask instruction:}
        Collect the pumpkin by holding left button.

        \vspace{5pt}

        \noindent\textbf{Target action sequence:}
        \[
        \mathbf a_t
        =
        \left[
        \langle\mathrm{BOA}\rangle,
        \langle\mathrm{NG}\rangle,
        Attack,
        m_{x,b_{t,x}}^{(\mathrm{NG})},
        m_{y,b_{t,y}}^{(\mathrm{NG})},
        \langle\mathrm{EOA}\rangle
        \right].
        \]

        \vspace{5pt}
        \hrule
        \vspace{7pt}

        \noindent\textbf{GUI Interaction Example}
        \par\vspace{3pt}
        \hrule
        \vspace{5pt}

        \begin{center}
            \includegraphics[
                width=0.5\linewidth
            ]{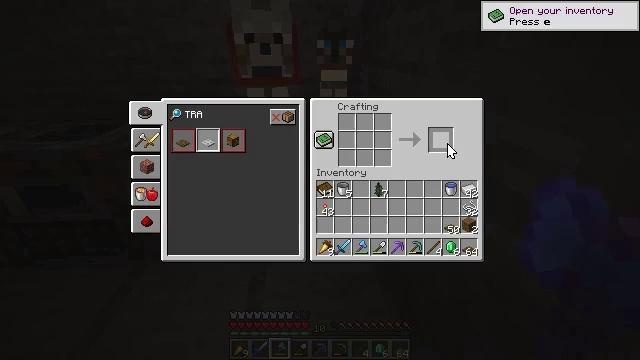}
        \end{center}

        \noindent\textbf{Subtask instruction:}
        Craft the fishing rod at the crafting table. 

        \vspace{5pt}

        \noindent\textbf{Target action sequence:}
        \[
        \mathbf a_t
        =
        \left[
        \langle\mathrm{BOA}\rangle,
        \langle\mathrm{GUI}\rangle,
        Use,
        m_{x,b_{t,x}}^{(\mathrm{GUI})},
        m_{y,b_{t,y}}^{(\mathrm{GUI})},
        \langle\mathrm{EOA}\rangle
        \right].
        \]

    \end{minipage}%
    }
\end{table*}

\end{document}